\documentclass[11pt,a4paper]{article}
\usepackage[hyperref]{emnlp2020}
\usepackage{times}
\usepackage{latexsym}

\usepackage{amsmath}
\usepackage{graphicx}
\usepackage{booktabs}
\usepackage{multirow}
\usepackage[most]{tcolorbox}
\newtcolorbox{affirmbox}{colback=red!5!white,colframe=red!75!black}
\usepackage{varwidth}
\usepackage[labelsep=period]{caption}
\usepackage{xurl}
\usepackage{fancyhdr}

\fancypagestyle{3296}{\fancyhf{}\fancyfoot[C]{\normalsize{3296}\vskip 0.085in\footnotesize{\textit{Findings of the Association for Computational Linguistics: EMNLP 2020}, pages 3296-3315\\November 16 - 20, 2020. $\copyright$2020 Association for Computational Linguistics}}}
\fancypagestyle{3297}{\fancyhf{}\fancyfoot[C]{\normalsize{3297}}}
\fancypagestyle{3298}{\fancyhf{}\fancyfoot[C]{\normalsize{3298}}}
\fancypagestyle{3299}{\fancyhf{}\fancyfoot[C]{\normalsize{3299}}}
\fancypagestyle{3300}{\fancyhf{}\fancyfoot[C]{\normalsize{3300}}}
\fancypagestyle{3301}{\fancyhf{}\fancyfoot[C]{\normalsize{3301}}}
\fancypagestyle{3302}{\fancyhf{}\fancyfoot[C]{\normalsize{3302}}}
\fancypagestyle{3303}{\fancyhf{}\fancyfoot[C]{\normalsize{3303}}}
\fancypagestyle{3304}{\fancyhf{}\fancyfoot[C]{\normalsize{3304}}}
\fancypagestyle{3305}{\fancyhf{}\fancyfoot[C]{\normalsize{3305}}}
\fancypagestyle{3306}{\fancyhf{}\fancyfoot[C]{\normalsize{3306}}}
\fancypagestyle{3307}{\fancyhf{}\fancyfoot[C]{\normalsize{3307}}}
\fancypagestyle{3308}{\fancyhf{}\fancyfoot[C]{\normalsize{3308}}}
\fancypagestyle{3309}{\fancyhf{}\fancyfoot[C]{\normalsize{3309}}}
\fancypagestyle{3310}{\fancyhf{}\fancyfoot[C]{\normalsize{3310}}}
\fancypagestyle{3311}{\fancyhf{}\fancyfoot[C]{\normalsize{3311}}}
\fancypagestyle{3312}{\fancyhf{}\fancyfoot[C]{\normalsize{3312}}}
\fancypagestyle{3313}{\fancyhf{}\fancyfoot[C]{\normalsize{3313}}}
\fancypagestyle{3314}{\fancyhf{}\fancyfoot[C]{\normalsize{3314}}}
\fancypagestyle{3315}{\fancyhf{}\fancyfoot[C]{\normalsize{3315}}}

\usepackage{microtype}

\aclfinalcopy 
\def\aclpaperid{2439} 

\newif\ifcomment
\commenttrue

\title{Detecting Stance in Media on Global Warming}

\author{Yiwei Luo$^1$ \qquad
  Dallas Card$^2$ \qquad
  Dan Jurafsky$^{1,2}$ \\
  Stanford University \\ 
  $^1$Department of Linguistics \hspace{15pt} $^2$Department of Computer Science \\
  \texttt{\{yiweil, dcard, jurafsky\}@stanford.edu}}

\begin{document}
\maketitle
\thispagestyle{3296}
\begin{abstract}
Citing opinions is a powerful yet understudied strategy in argumentation. For example, an environmental activist might say, ``Leading scientists agree that global warming is a serious concern,'' framing a clause which affirms their own stance (\textit{that global warming is serious}) as an opinion endorsed (\textit{[scientists] agree}) by a reputable source (\textit{leading}). In contrast, a global warming denier might frame the same clause as the opinion of an \textit{untrustworthy} source with a predicate connoting \textit{doubt}: ``\textit{Mistaken} scientists \textit{claim} [...].'' Our work studies opinion-framing in the global warming (GW) debate,\footnote{Throughout, we use the term \textit{debate} to refer to the existence of contrasting opinions about GW expressed in the media; it is important to emphasize that there is virtually 100\% consensus among scientists regarding the reality of anthropogenic global warming \citep{powell2017scientists}.} an increasingly partisan issue that has received little attention in NLP. We introduce \textbf{Global Warming Stance Dataset (GWSD)}, a dataset of stance-labeled GW sentences, and train a BERT classifier to study novel aspects of argumentation in how different sides of a debate represent their own and each other's opinions. From 56K news articles, we find that similar linguistic devices for self-affirming and opponent-doubting discourse are used across GW-accepting and skeptic media, though GW-skeptical media shows more opponent-doubt. We also find that authors often characterize sources as hypocritical, by ascribing opinions expressing the author's \textit{own} view to source entities known to publicly endorse the \textit{opposing} view.
We release our stance dataset, model, and lexicons of framing devices for future work on opinion-framing and the automatic detection of GW stance.
\end{abstract}

\section{Introduction}
\begin{figure}
    \centering
    \includegraphics[scale=0.46]{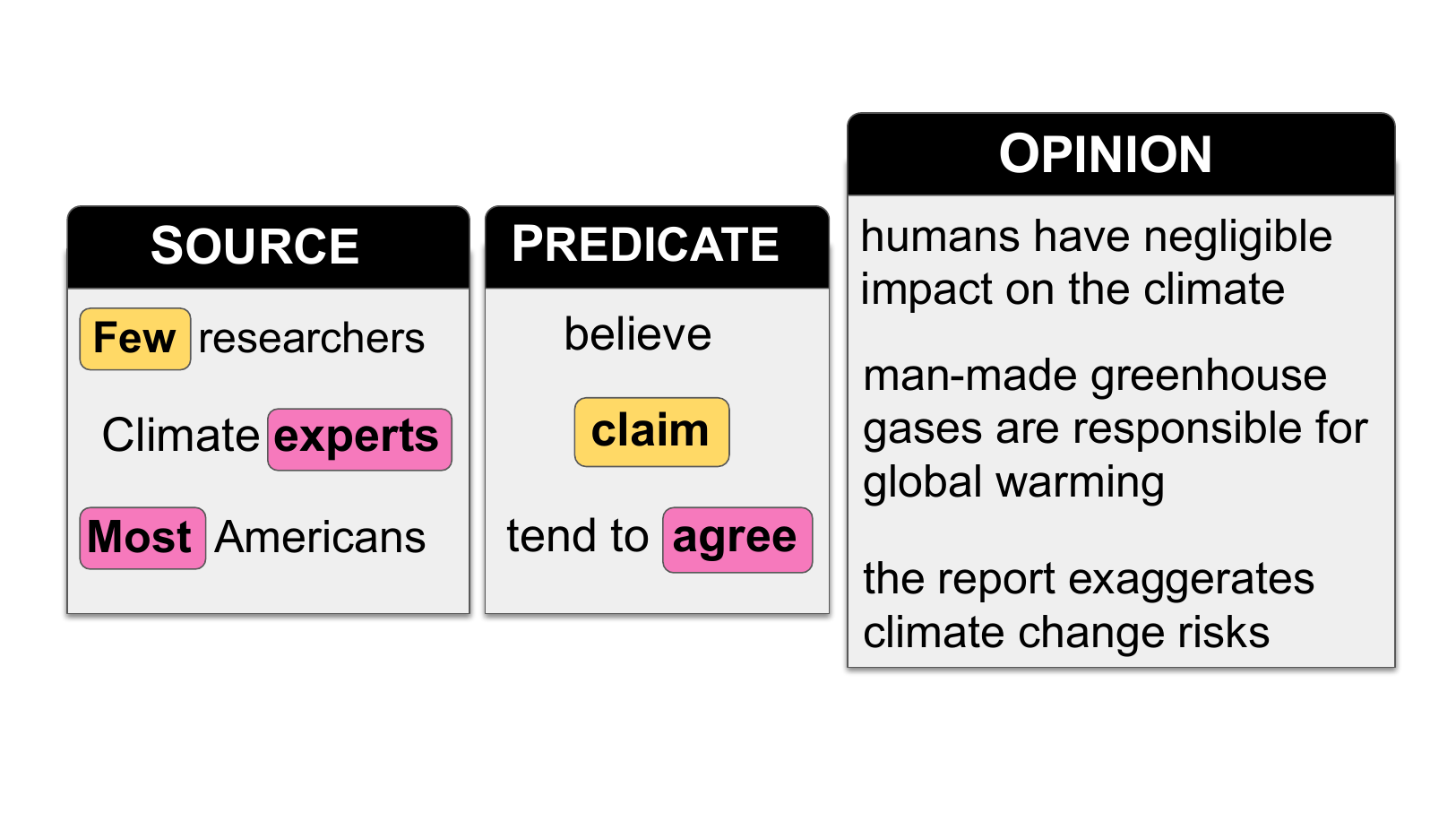}
    \caption{Examples of \textsc{source}, \textsc{predicate}, and \textsc{opinion} components, and within components, examples of
    \begin{tcolorbox}[hbox,tcbox raise base,colback=magenta!70,boxrule=1.1pt,top=0.03pt,bottom=0.03pt,left=0.03pt,right=0.03pt,nobeforeafter]
    \textbf{affirming}
    \end{tcolorbox} and \begin{tcolorbox}[hbox,tcbox raise base,colback=yellow!70!white,boxrule=1.1pt,top=0.03pt,bottom=0.03pt,left=0.03pt,right=0.03pt,nobeforeafter]
    \textbf{doubting}
    \end{tcolorbox} framing devices.} 
    \label{fig:slot_examples}
\end{figure}

Ascribing opinions to other people is a powerful yet understudied strategy in argumentation. For example, an environmental activist might say, ``\textbf{\emph{Leading}} scientists \textbf{\emph{agree}} that global warming is serious,'' whereas a global warming denier could say, ``\textbf{\emph{Mistaken}} scientists \textbf{\emph{claim}} that global warming is serious.'' In both these examples, the embedded clause (\textit{that global warming is serious}) is presented as an opinion belonging to a source entity (\textit{scientists}). However, differences in the choice of predicate (\textit{agree} vs. \textit{claim}) and in how the source is described lead to very different interpretations. We henceforth refer to the use of such [ENTITY] [EXPRESS] [STATEMENT] sentences as \textbf{opinion-framing}, and to the respective components as the \textsc{source}, \textsc{predicate}, and \textsc{opinion} (see Fig. \ref{fig:slot_examples}).

Despite its pervasiveness in argumentative discourse, opinion-framing is understudied as a persuasive strategy. This paper studies opinion-framing in the media coverage of global warming (GW), an increasingly partisan issue in the United States \citep{pewclimate} that has received little attention in NLP despite its real world urgency. We focus on acts of opinion-framing representing \textit{self-affirming} and \textit{opponent-doubting} discourses, i.e., discourse affirming one's own \textsc{opinions} (embedded clauses ascribed to a \textsc{source}, as depicted in Fig. \ref{fig:slot_examples}) and discourse casting doubt on the other side's. Studying such discourses requires a way to identify the stance of a given \textsc{opinion} with respect to GW, but this is a challenging task.

\thispagestyle{3297}
To this end, we introduce \textbf{GWSD: G}lobal \textbf{W}arming \textbf{S}tance \textbf{D}ataset, a dataset for detecting and analyzing GW stance in text. We collect human judgments of GW stance for 2K sentences with Amazon Mechanical Turk (AMT)\footnote{We also experimented with tweets from GW-activists/skeptics and headlines from extreme conservative/liberal outlets as potential sources of softly stance-labeled sentences, but found that classifiers trained on these data perform poorly on news discourse.}  and use our dataset to train a BERT-based classifier that achieves 75\% accuracy (competitive with human performance) for GW stance detection. Extending prior work in NLP and linguistics, we develop lexicons of affirming and doubting framing devices with respect to the \textsc{predicate} that embeds the \textsc{opinion} (e.g., \textit{know} vs. \textit{claim}) and the \textsc{source} to which the opinion is ascribed (e.g., \textit{a peer-reviewed study} vs. \textit{a misleading paper}) (see Fig. \ref{fig:slot_examples}). 

We then apply our model and lexicons to study two questions about opinion-framing in argumentation: \textbf{Q1:} Do different sides of a debate (in this case, GW-accepting and GW-skeptical media) show symmetry in their use of self-affirming and opponent-doubting discourse? We might expect some similarities (e.g., the use of \textit{agree} to frame \textsc{opinions} expressing one's own side's stance, or the use of \textit{claim} to cast doubt on \textsc{opinions} from the opposing side), but given inherent asymmetries in the nature of the GW debate, it is not clear whether such strategies will be found across sides to equal extents.

Second, since opinion-framing is a way of putting words into someone's mouth, we also ask \textbf{Q2:} In cases where \textsc{opinions} are ascribed to a named entity with a known (public) stance, does the stance of the \textsc{opinion} match the expected stance of the named entity?

Applying our model to a set of 500K \textsc{opinions} (\textbf{Op$_{full}$}) extracted from 56K GW articles, we find that GW-skeptical media engages in comparatively more opponent-doubt, though both sides of the debate show more self-affirmation overall, and use similar sets of framing devices for each respective discourse type. We also find that opinion-framing does indeed ascribe \textsc{opinions} differing from the overt views of entities to those entities nonetheless, as part of a rhetorical strategy of ascribing hypocrisy:
authors portray their \textit{own} \textsc{opinion} as being held (in private) by figures who endorse the \textit{opposite} \textsc{opinion} (in public).

Our contributions are the following:
\begin{enumerate}
    \item \textbf{GWSD}, a dataset of 2K sentences from GW news with annotations for stance.
    \item A weighted extension of BERT competitive with human performance for classifying the stance of a sentence with respect to GW.
    \item Lexicons of affirming and doubting \textsc{predicates} (\textit{e.g., know, claim}) and \textsc{source} modifiers (\textit{e.g., peer-reviewed, misleading}).
    \item Analyses on a set of 500K opinions from GW news to illustrate the utility of our dataset and lexicons for studying opinion-framing.
\end{enumerate}
We release our dataset, model, and lexicons as part of this paper.\footnote{\textcolor{blue}{\url{https://github.com/yiweiluo/GWStance}}}

\section{Related work}

Our work is related to social psychology research on persuasion  \citep{cialdini1993influence,orji2015gender} and recent NLP research on argumentation, such as predicting argument convincingness \citep{habernal-gurevych-2016-argument,simpson-gurevych-2018-finding} and studying discourse-level and non-linguistic features predictive of persuasion \citep{yang2017persuading,zhang2016conversational}. The latter's work on self- vs. opponent-coverage is particularly relevant to the GW debate and we apply a similar categorization to the stance of ascribed opinions.

Also relevant is the literature on factuality and speaker commitment \citep{de2011veridicality,soni2014modeling,werner2015committed,rudinger-etal-2018-neural-models,jiang-de-marneffe-2019-know}, and relatedly, work studying how words can express subjectivity or bias \citep{riloff2003learning,recasens2013linguistic,pryzant2020automatically}. Our current paper builds upon previous work by examining such triggers as opinion-framing devices in an argumentation context, where biases related to people's prior beliefs may interact with the lexical effects of these words.

Opinion-framing can be thought of as a special case of the broader phenomenon of framing as discussed in the communications and political science literatures \citep{entman1993framing, lakoff2006framing,chong2007framing}, as well as in NLP \citep{tsur.2015,field.2018,roy2020weakly}. Both phenomena serve to emphasize particular aspects of an issue, and are often used with the intent to influence perception of that issue. Our attention to the component of \textsc{source} in instances of opinion-framing is also informed by communications research on the \textit{messenger effect} (that people's perceptions of a message may depend heavily on the message source) \citep{bolsen2019impact,myrick2020pope,fielding2020using,esposo2013shooting}. Furthermore, our interest in predicates of opinion attribution is inspired by communications studies examining how the choice of predicate (\textit{say} vs. \textit{assert}) can encode journalist stance \citep{caldas2002reporting} and bias audience perception of the quoted entity \citep{gidengil2003talking}. Finally, our dataset contribution builds on Mohammad et al. \citeyearpar{mohammad2016semeval}, who created the first climate change stance task and dataset.

\section{GWSD: A dataset for GW stance}
\label{sec:dataset} 
\thispagestyle{3298}
To enable our study of opinion-framing, and to facilitate further work on stance, we create a new publicly-available dataset of 
\textsc{opinion} spans extracted from GW news articles (described in \S\ref{ssec:news articles}) that we have annotated with stance judgements using AMT (\S\ref{ssec: mturk}). 
To investigate potential annotator biases, we study the impact of annotator characteristics on their perception of stance (with approval from our Institutional Review Board) (\S\ref{ssec:dem effects}), and combine ratings so as to infer a distribution over stance labels for each span while accounting for bias (\S\ref{ssec:aggregation}), which we release along with the raw annotations.

\subsection{Extracting sentences for the dataset}
\label{ssec:news articles} 
\begin{figure*}
    \centering
    \includegraphics[scale=0.25]{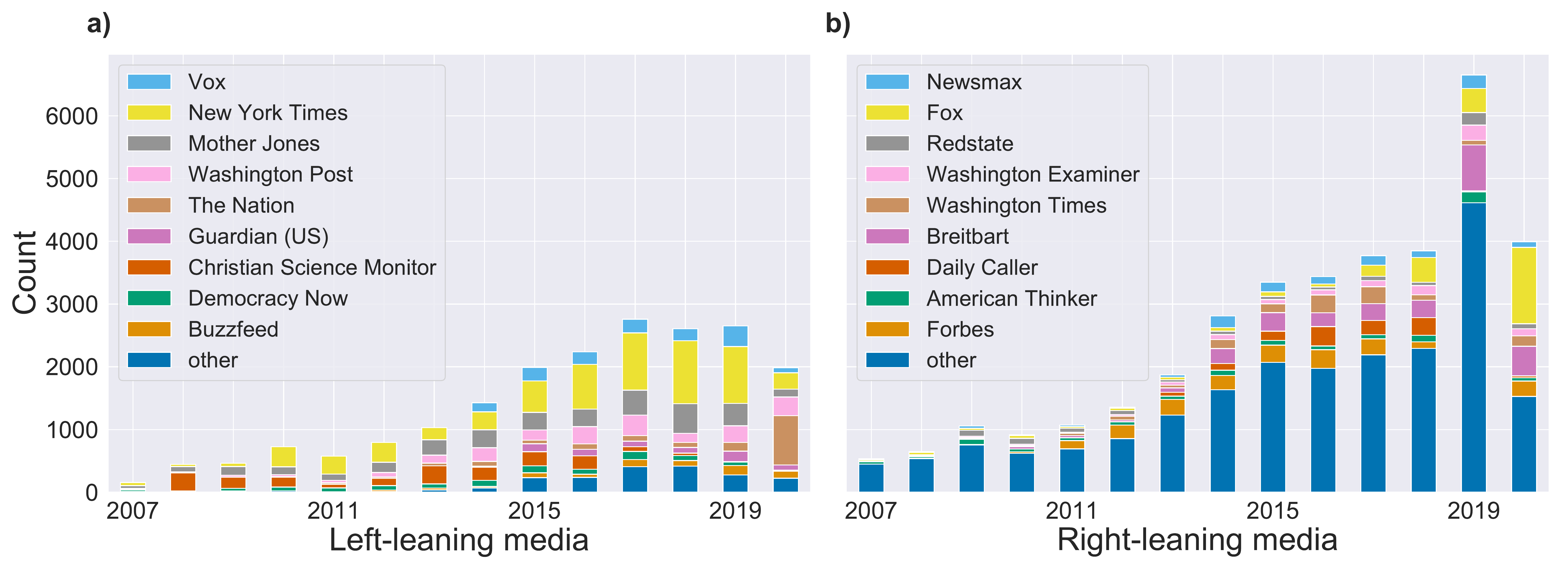}
    \caption{Number of GW articles in our dataset from 2007 to 2020 in \textbf{a)} Left-,  \textbf{b)} Right-leaning media.}
    \label{fig:articles over time}
\end{figure*}

Our base dataset consists of \textsc{opinion} spans extracted from 56K GW news articles, published from Jan. 1, 2000 to April 12, 2020 by 63 U.S. news sources. We collected these articles using the MediaCloud API\footnote{\url{https:/ /cyber.harvard.edu/research/mediacloud}} and SerpAPI.\footnote{\url{https:/ /serpapi.com/search-api}} The keywords we used for API requests were: \{climate change, global warming, fossil fuels, carbon dioxide, methane, co2\}. We note that some of the articles in our dataset come from newswires (N=1.3K), but as we show later, including wire articles does not affect our studies' conclusions. Moreover, since it is ultimately up to media outlets to decide which wire articles to publish, we believe that instances of opinion-framing from wire articles are still reflective of what an outlet endorses (despite not originating from the outlet). We also include op-ed articles in our dataset, as their exclusion is made challenging by idiosyncrasy in their coding across outlets. Future work might exclude op-ed articles for model training and analysis. Please refer to Appendix \ref{sec:data collection details} for details on our filtering and de-duplication steps. Tab. \ref{tab:cc outlet stats} and Fig. \ref{fig:articles over time} summarize the distribution of articles by source.

\begin{table}[]
    \centering
    \begin{tabular}{c c c c } \toprule
         \multicolumn{2}{c}{\textbf{Left-leaning outlets}} & \multicolumn{2}{c}{\textbf{Right-leaning outlets}} \\ \midrule
         NYT & 6K & Breitbart & 2.7K \\
         Moth. Jones & 3.2K & Fox & 2.6K \\
         WaPo & 2K & Forbes & 2K \\
         CS Monitor & 1.9K & Wash. Times & 1.4K \\
         The Nation & 1.4K & Daily Caller & 1.2K \\
         Vox & 1.4K & Newsmax & 1.2K \\
         Dem. Now & 1K & Wash. Exam. & 1K \\ \midrule 
         \textit{Total} & 20K & \textit{Total} & 36K \\ \bottomrule
    \end{tabular}
    \caption{Number of unique articles from the top 7 
    left-leaning and right-leaning
    media outlets in our dataset (LL and RL), 
    by volume of articles contributed. We categorize political leaning using the Media Bias/Fact Check project.}
    \label{tab:cc outlet stats}
\end{table}

To identify the rhetorical components of relevant sentences, we make use of syntactic dependency parsing to extract
embedded \textsc{opinion} spans (e.g., \textit{Scientists believe that [\textbf{climate change requires immediate action}]}) from a given article, 
as well as
spans 
for
\textsc{source} (who or what the \textsc{opinion} is ascribed to) and \textsc{predicate} (the verb that syntactically embeds the \textsc{opinion}).
Note that we exclude \textsc{opinions} under the scope of negation or modals.

Our pipeline consists of first passing each article through the spaCy pre-processing pipeline with a neural coreference resolution add-on,\footnote{\url{https://github.com/huggingface/neuralcoref}, which implements the model from Clark and Manning \citeyearpar{clark-manning-2015-entity}.} then extracting and annotating instances of \textsc{source, predicate} and \textsc{opinion} using a rule-based algorithm (please refer to Appendix \ref{sec:appendix spacy pipe}). To validate our algorithm, we manually annotated 25 articles and compared results. We found that a dependency parsing-based approach has a high recall, identifying all clausal complements including some false positives such as indirect questions and subjunctive clauses. We therefore used several lexical resources to filter the extracted clauses to indicative statements.

Finally, since many of the \textsc{opinions} that we extracted are not explicitly on the topic of GW, we only keep the \textsc{opinion} spans that contain a stem from a set of 73 manually curated keywords (e.g., \textit{climat, environ, temperatur}). 

\subsection{Crowd-sourcing labels for the dataset}
\label{ssec: mturk}
\thispagestyle{3299}
We used AMT to label a subset of 2,050 \textsc{opinion} spans containing high-precision keywords (see Appendix \ref{app:prec-keywords}). The set of 2,050 spans was constructed iteratively by randomly sampling, then manually filtering spans containing potentially upsetting material (e.g., mocking Greta Thunberg’s disability) or that were off-topic (e.g., used ``climate'' in the sense of a workplace environment). For each \textsc{opinion}, we collected judgements as to whether it expresses the target opinion: ``Climate change/global warming is a serious concern,'' with the potential labels being ``agree,'' ``neutral,'' or ``disagree.''

Following 4 pilot studies, we decided to collect 8 judgements per item (to enable robust analysis of demographic variation in annotator judgements), for a total of 16,400 annotations, paying the California minimum wage of \$12USD per hour. Using typical exclusion criteria, we recruited a set of 398 qualified annotators over 5 rounds and had them rate 30-50 items. We also asked for basic demographic information and their personal opinions on a series of questions related to GW (see Appendix \ref{app:appendix example mturk} for details and an example).

Although stance datasets are typically created with the notion of a ``true'' label for each item, we note that there is some degree of inherent ambiguity in this task due to the complex nature of the GW debate as well as the items' being taken out of context. The average inter-annotator agreement (IAA) measured as Krippendorff's alpha ranged from 0.54 to 0.64 over the 5 rounds of annotation, though the vast majority of disagreements were between adjacent labels. Some items with high disagreement are shown in Tab. \ref{tab:hard items}, showing the possibility of genuine ambiguity in GW stance. 

\begin{table*}[ht]
    \centering
    \begin{tabular}{l} \toprule
         \textbf{1.} Global warming is inevitably going to be, at best, managed. \textbf{2.} Global warning will be over- \\ ridden by
         this effect, giving humankind and the Earth 30 years to sort out our pollution. \\ 
         \textbf{3.} The global warming debate is over. \textbf{5.} Global warming would open stretches of the Arctic \\ Ocean to shipping and drilling. \\\bottomrule
    \end{tabular}
    \caption{Examples of items eliciting the highest disagreement among annotators (measured as entropy over labels). Each of these items was annotated with all 3 labels -- ``agree,'' ``neutral,'' and ``disagree.'' The stance of these items seems to depend not only on the linguistic content present but also on who the speaker might be, or what the statement is said in response to, making them difficult to label.}
    \label{tab:hard items}
\end{table*}

\subsection{Demographic effects on annotation}
\label{ssec:dem effects}
Given that GW has become a polarized issue in the US, we test whether we observe any bias related to party affiliation in stance annotation.
Past work has called attention to the importance of considering demographic biases in annotation  \citep{cowan.2003,sap.2019}. Intuitively, we might expect that those skeptical of GW would be more likely to perceive a sentence as exaggerating its threat, and therefore more likely to classify the sentence as one that suggests that GW is a serious concern (even though they themselves may disagree).  

In order to test for the presence of demographic bias, we make use of Bayesian hierarchical ordinal regression models to estimate the effect of various annotator characteristics, such as party affiliation \citep{gelman.hill.2007}, which we fit using Stan \citep{carpenter.2017}. Because we have 8 annotations per item and 30-50 annotations from each annotator, we model variation in both items and worker biases, with the latter drawn from a hierarchial prior incorporating annotator characteristics (please see Appendix \ref{app:demographic} for details).

As expected, we do find clear evidence of a slight bias along party lines. For a typical \textsc{opinion}, (self-identified) Republicans are approximately 1.05 ($\pm$ 0.016 s.d.) times more likely to label an item as ``agree'' compared to non-Republicans, and similarly less likely to respond with ``disagree.'' We see the opposite trend for Democrats, though the effect of the latter is mitigated by the inclusion of additional covariates. More surprisingly, we also find a slight gender bias, with those who self-identify as female being 1.04 times more likely to respond with ``agree''  ($\pm$ 0.011 s.d.). This effect is robust to the inclusion of other variables, but should be interpreted with caution, as women were somewhat underrepresented in our study (see Tab. \ref{tab:agree_table} in Appendix \ref{app:demographic} for full modeling results). Regardless, this reinforces the importance of taking potential annotator biases into account \citep{cowan.2003,sap.2019} and is suggestive for further research. 

\subsection{Aggregating annotations}
\label{ssec:aggregation}

Because some workers are more reliable than others, we again make use of Bayesian modeling to aggregate the annotations for each item. Drawing inspiration from MACE \citep{hovy-etal-2013-learning},
we fit a model which includes a distribution over labels associated with each item (i.e., agree, neutral, disagree), corresponding biases for each annotator, and a parameter indicating the degree to which they are influenced by their own biases. Whereas MACE assumes that annotators sometimes choose labels at random on individual instances, but otherwise identify the true label, we assume that annotators are always somewhat influenced by their biases, but to differing degrees.
This model allows us to simultaneously infer a distribution over labels for each instance (i.e., the probability of each label being chosen by a typical worker), as well as bias and vigilance terms for each annotator.
(Please see Appendix \ref{app:aggregating} for full model details). Based on this model, we assign the highest probability label to each \textsc{opinion}, as summarized in Table \ref{tab:label_stats}.

\begin{table}[ht]
    \centering
    \begin{tabular}{c c} \toprule
         Label & Count \\ \midrule
         neutral & 873 \\
         agree & 777 \\
         disagree & 400 \\ \bottomrule
    \end{tabular}
    \caption{Distribution of labels in \textbf{GWSD}, as aggregated by our model when the label with highest inferred probability is selected.}
    \label{tab:label_stats}
\end{table}
\thispagestyle{3300}
\section{A model for GW stance classification}
\label{ssec:classifier training}

In order to classify stance in \textbf{\textit{Op$_{full}$}}, the full dataset of 500K \textsc{opinions}, 
we train a model using the
set of 2K annotated examples.
The goal of 
this task 
is to predict the stance of a sentence $S$
toward the target opinion $T$ (``Climate change/global warming is a serious concern''). 
To evaluate performance, we first select a random test set of 200 annotated instances (stratified by label and political leaning of the source media outlet) and use 5-fold cross validation to train on the remaining 1850 examples.

Here,
we report on variations on a BERT classifier \citep{devlin-etal-2019-bert}, as well as a linear baseline, in order to provide a sense of relative performance in comparison to past work.
To ensure comparison against a strong baseline, we perform a grid search over hyperparameters
for both approaches, and choose the best model from each according to validation accuracy, 
evaluating only the best model of each type on the held-out test set. 

For our neural model, 
we 
use the general-purpose 
BERT$_{base}$  architecture, trained by minimizing cross-entropy loss. We use the Transformers library\footnote{\url{https://huggingface.co/transformers/}} as the basis for the models that we develop and compare. As potential augmentations, we experiment with a) fine-tuning the base model as a language model to unlabeled data; b) including the text of the target opinion as an input to the model; and c) using label weights as opposed to simply using the most probable label. For the weighted version, we include a copy of each training instance with each label, along with an instance weight corresponding to the label probability estimated by our 
label aggregation
model
above.
(Full details of hyperparameter tuning in Appendix \ref{app:hyper}). 

The test-set performances of best models we obtain are shown in Table \ref{tab:classifiers}, along with majority class and human performance (see Appendix \ref{app:aggregating}).
The best performing BERT model used weighted data and incorporated the target opinion as an input, but was not fine-tuned as a language model. The accuracy of this model is competitive with human performance (estimated using leave-one-out subsets of 10\% of annotators), and mis-classifications of ``agree'' as ``disagree'' or vice versa occurred in less than 9\% of test examples.

Further inspection of the validation results reveals that training on the weighted data offers a statistically significant improvement on validation accuracy, but the expected performance is statistically indistinguishable with respect to fine-tuning and/or incorporating the target opinion as an input. The best linear model was a simple $l_2$-weighted logistic regression classifier using unigrams and bigrams (details in Appendix \ref{app:hyper}).

\begin{table}[ht]
    \centering
    \small
    \begin{tabular}{l c c c c c} \toprule
         & acc & $F_{\texttt{A}}$ & $F_{\texttt{N}}$ & $F_{\texttt{D}}$ & $F_{\texttt{avg}}$ \\ \midrule
         \textit{Majority class} & 0.43 & 0.0 & 0.52 & 0.0 & 0.17 \\
        \textit{Linear} & 0.62 & 0.55 & 0.66  & 0.56 & 0.60 \\
        \textit{BERT} & \bf{0.75} & \bf{0.68} & \bf{0.76} & \bf{0.75} & \bf{0.73} \\
        \textit{Human} & 0.71 & & & & \\ \bottomrule
    \end{tabular}
    \caption{Test-set performance, reported as accuracy, and macro-F1 score for each label (agrees, neutral, disagrees) and on average, of the best model of each type, trained using hyperparameters values corresponding to the model with the best cross-fold validation performance, with the overall best performing model shown in bold. See Appendix \ref{app:aggregating} for further details on how human performance was estimated.}
    \label{tab:classifiers}
\end{table}
\thispagestyle{3301}
\section{Analyses}
In this section, we first describe the lexicons of framing devices we use for our analyses (\S\ref{ssec:framing devices}). We then present analyses that address our two research questions. 

In \S\ref{ssec:study 1}, we find that \textit{qualitatively}-speaking, both sides leverage similar linguistic framing devices for self-affirmation and opponent-doubt, but \textit{quantitatively}-speaking, GW-skeptical media engages in more opponent-doubt. In \S\ref{ssec:study 2}, we find that both sides use opinion-framing to ascribe \textsc{opinions} expressing their \textit{own} stance to \textsc{sources} known to publicly endorse the \textit{opposing} view, thereby depicting such \textsc{sources} as hypocritical.

\subsection{Linguistic framing devices}
\label{ssec:framing devices}
Since GW opinion is closely connected to one's attitude toward scientific evidence, we focus on framing devices with epistemic and evidential connotations in creating lexicons of affirming and doubting framing devices. We draw from work on factuality, commitment, and persuasion, as well as our own lexical semantic analysis, to create seed word sets; these seed sets are then augmented using WordNet to become our final lexicons.

\paragraph{Affirming devices} We include factive and semi-factive predicates (\textit{point out, understand} (N=20)), studied extensively in de Marneffe et al. \citeyearpar{de2011veridicality}, Saur{\'i} and Pustejovsky \citeyearpar{sauri-pustejovsky-2012-sure}, Rudinger et al. \citeyearpar{rudinger-etal-2018-neural-models}, Jiang and de Marneffe \citeyearpar{jiang-de-marneffe-2019-know}, Ross and Pavlick \citeyearpar{ross2019well}, among others. We add verbs with connotations of factivity and/or high subject commitment (\textit{confirm, attest, certify, validate} (N=7)). We also add high commitment adjectives (\textit{proven, settled} (N=4)) and
adjectives of ``hyping'' from Lerchenmueller et al. \citeyearpar{lerchenmueller2019gender} (\textit{breakthrough, expert} (N=38)). To complement these adjectives that affirm the \textit{quality} of evidence, we add modifiers that affirm the \textit{quantity} of evidence and index consensus (\textit{many, numerous, dozens of} (N=11)).

\paragraph{Doubting devices} We include words from semantic fields largely antonymous to those represented in the affirming seed words: neg-factive verbs \citep{sauri2009factbank} such as \textit{claim, pretend} (N=5), low commitment verbs (\textit{doubt, dispute} (N=3)), low commitment adjectives (\textit{dubious, so-called} (N=7)),  adjectives of undermining (\textit{flawed, debunked} (N=47)) and adjectives indexing lack of consensus (\textit{few, contentious} (N=6)). We additionally include verbs with argumentative connotations (\textit{argue, insist} (N=11)), as these can reinforce frames of debate and controversy. 

We hope that our full lexicons (see Appendix \ref{app:framing devices}) will be useful for future work that looks at opinion-framing, especially in the context of other scientific debates (e.g., the COVID-19 pandemic).

\subsection{Study 1 results}
\label{ssec:study 1}
We apply our stance classification model to \textit{\textbf{Op$_{full}$}}\footnote{Because \textsc{opinion} spans from certain media outlets are over-represented in \textit{\textbf{Op$_{full}$}}, we repeat all analyses in Studies 1 and 2 while excluding data points from the top 5 LL and RL outlets (10 total) and obtain largely similar results (see Appendix \ref{app:subsampled results}) to those presented in the main paper.} to get a stance label for all embedded \textsc{opinions}. We restrict our analysis to \textsc{opinions} receiving a non-neutral label, as we can better guarantee having few mis-classifications of GW-agree (the sentence agrees with the target that GW is a serious concern) as GW-disagree (the sentence disagrees with the target that GW is a serious concern), and vice versa. We use political leaning as categorized by the Media Bias/Fact Check project\footnote{\url{https://mediabiasfactcheck.com/}} as a proxy for stance toward GW, with left-leaning and right-leaning outlets (LL and RL) corresponding to GW-accepting and GW-skeptical media, respectively. To find instances of self-affirmation in GW-accepting media, we retrieve GW-agree \textsc{opinions} occurring with a \textsc{predicate} or \textsc{source} modifier from the group of affirming devices (e.g., \textit{show, peer-reviewed}); to find instances of opponent-doubt, we retrieve GW-disagree \textsc{opinions} occurring with \textsc{predicates} or \textsc{source} modifiers from the set of doubting devices (e.g., \textit{claim, misleading}). This is repeated for GW-skeptical media, with \textsc{opinion} stances swapped.

\begin{figure}[ht]
    \centering
    \includegraphics[scale=0.26]{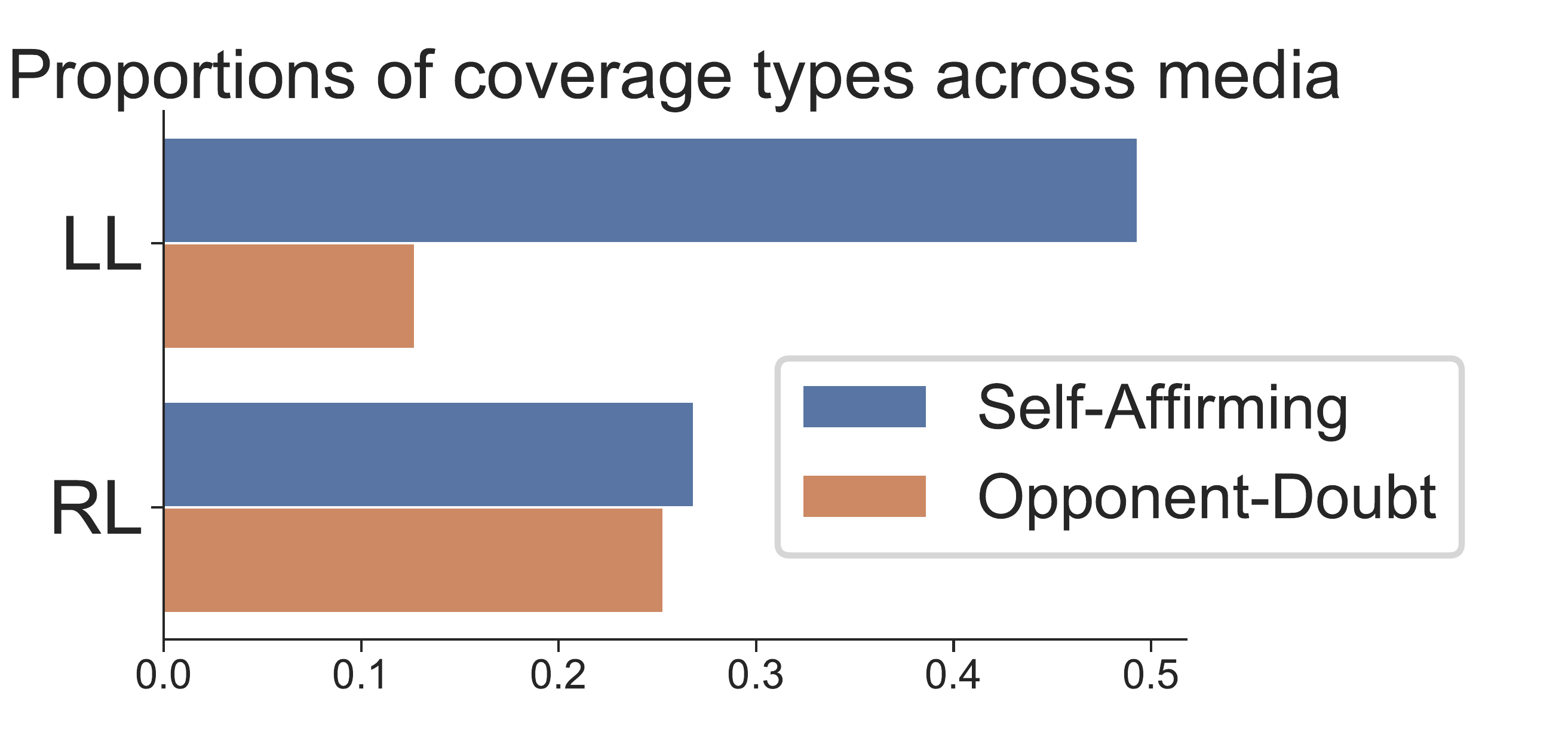}
    \caption{Proportions (among non-neutral \textsc{opinions}) of self-affirming vs. opponent-doubting coverage in LL and RL, showing that LL primarily exhibits discourse where a GW-agree \textsc{opinion} occurs with an affirming device, whereas RL exhibits more balanced amounts of self-affirmation and opponent-doubt. Most of the remaining \textsc{opinions} are framed by words beyond those in our lexicons.} 
    \label{fig:coverage_barplots}
\end{figure}

The resulting distribution over coverage types is shown in Fig. \ref{fig:coverage_barplots}, indicating that the two sides are \textit{not} symmetric in terms of their quantities of each coverage type: though both sides engage in more self-affirmation overall, GW-skeptical media (i.e., RL) shows a greater amount of opponent-doubt. This pattern corroborates prior work documenting the use of doubt by opponents of GW to dilute the scientific consensus \citep{oreskes2011merchants}.

\begin{figure}[ht]
    \centering
    \includegraphics[scale=0.36]{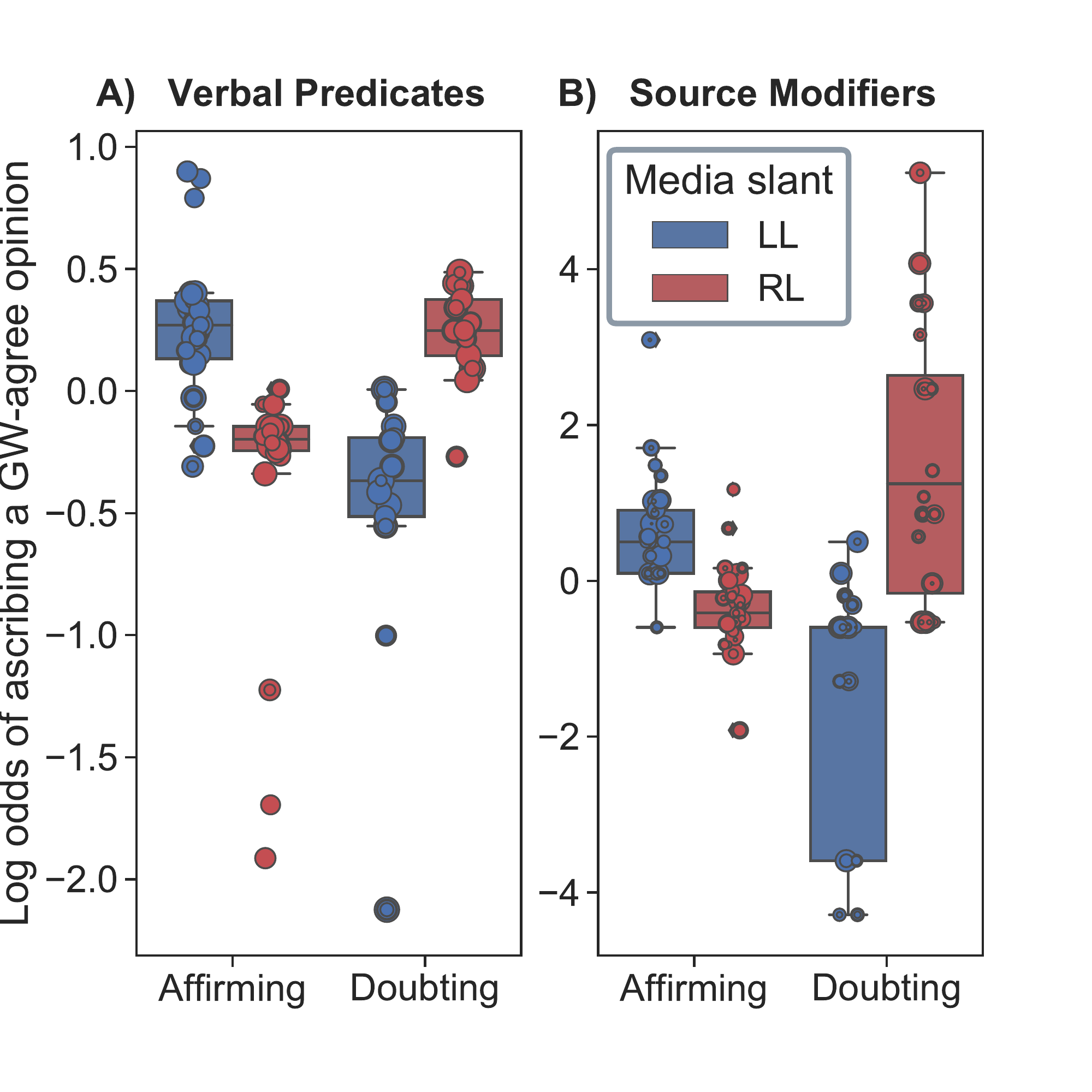}
    \caption{Distribution of the (log) odds of ascribing a GW-agree \textsc{opinion} in LL and RL for affirming and doubting \textbf{a)} \textsc{predicates}; \textbf{b)} \textsc{source} modifiers, showing that LL tends to ascribe GW-agree \textsc{opinions} using affirming devices over doubting devices, whereas RL tends to ascribe GW-agree \textsc{opinions} using doubting over affirming devices. Each point represents one framing device, and the size corresponds to its frequency in \textit{\textbf{Op$_{full}$}}.}
    \label{fig:pred_mod_boxplots}
\end{figure}

\begin{figure}[ht]
    \centering
    \includegraphics[scale=0.25]{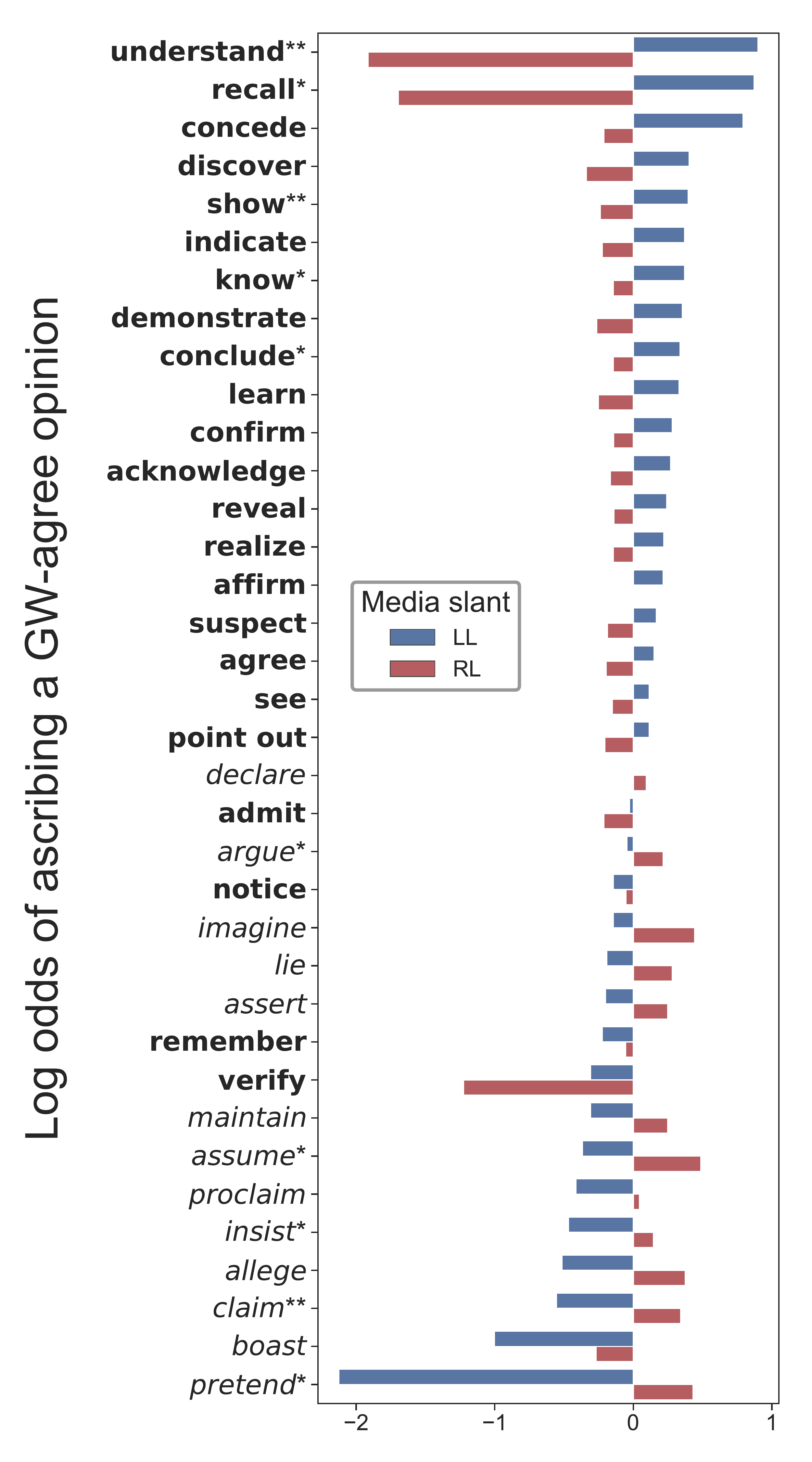}
    \caption{Log odds of ascribing a GW-agree \textsc{opinion} for \textbf{affirming} and \textit{doubting} predicates present in LL and RL, showing an overall symmetry in the devices LL and RL use for self-affirmation and opponent-doubt. A double asterisk (**) indicates a significant bias for GW-agree \textsc{opinions} in both LL and RL; (*) indicates significance in one side. Significance ($p<0.05$) is determined via a chi-squared test and applying Benjamini-Hochberg correction with a false discovery rate of 0.1. Word order is given in descending value of log odds, as measured in LL.}
    \label{fig:all_pred_bars}
\end{figure}

\begin{figure}[ht]
    \centering
    \includegraphics[scale=0.25]{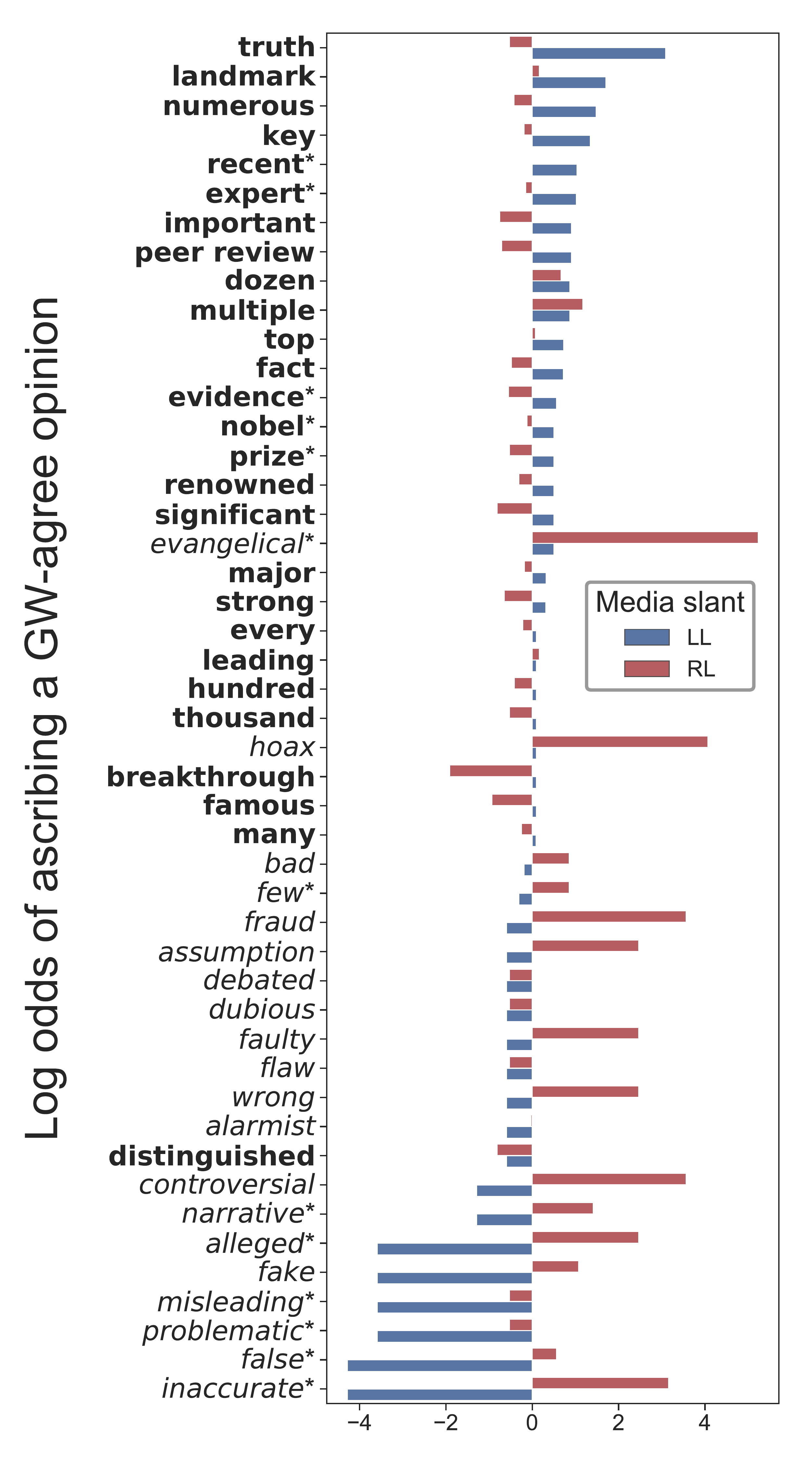}
    \caption{Log odds of ascribing a GW-agree \textsc{opinion} for the \textbf{affirming} and \textit{doubting} modifiers present in LL and RL. 
    An asterisk (*) indicates a significant bias for GW-agree \textsc{opinions} in either LL or RL. Significance ($p<0.05$) is determined via a chi-squared test and applying Benjamini-Hochberg correction with a false discovery rate of 0.1. Word order is given in descending value of log odds, as measured in LL.}
    \label{fig:all_mod_bars}
\end{figure}

\thispagestyle{3302}
Turning to qualitative aspects of self-affirming and opponent-doubting discourse, we find that the two sides show symmetry in the framing devices used: devices that LL tends to use to frame GW-agree \textsc{opinions} (e.g., \textit{understand, recall, discover}; \textit{important, peer review}) tend to be used by RL for GW-disagree \textsc{opinions}, and devices that RL uses to frame GW-agree \textsc{opinions} (e.g., \textit{pretend, claim; inaccurate, alleged}) tend to be used in LL for GW-disagree \textsc{opinions} (see Fig. \ref{fig:pred_mod_boxplots}). We measure the tendency for a framing device to occur with a given \textsc{opinion} stance as a log-odds-ratio between the number of times it frames \textsc{opinions} of each stance, excluding words that occur under 20 times (see Appendix \ref{app:log odds} for details). Broken down by the individual framing device (Figs. \ref{fig:all_pred_bars}-\ref{fig:all_mod_bars}), we also see that, with some exceptions, the use of framing devices across LL and RL displays some symmetry. Notably, there seems to be a lack of affirming modifiers framing GW-agree \textsc{opinions} in RL, suggesting that RL uses different modifiers to qualify \textsc{sources} as convincing.\footnote{As a robustness check, we repeat the same log-odds computation for the subset of data that excludes articles from newswires and find that the results are highly correlated with the full dataset (Pearson's \textit{r} = 0.90, $p<0.0001$ for verbs, Pearson's \textit{r} = 0.82, $p<0.0001$ for modifiers).} 

\subsection{Study 2 results}
\label{ssec:study 2}
How faithfully does the media ascribe \textsc{opinions} to \textsc{sources}? We use Wikipedia lists\footnote{Activist lists: \url{https://en.wikipedia.org/wiki/Category:Climate_activists}, \url{https://en.wikipedia.org/wiki/Category:Climate_change_environmentalists}. Unfortunately, the lists we used for climate change skeptics and deniers have since been deleted by Wikipedia. We manually removed entries that are neither people nor organizations, e.g., ``Environmental Activism of Al Gore.''} for GW-activist and GW-skeptic entities (Greta Thunberg, The Sierra Club; William Happer, The Heartland Institute) to label the stance of \textsc{sources} that are named entities, after using fuzzy matching to resolve \textsc{sources} to a canonical form. We define an \textsc{opinion} as \textit{faithfully} ascribed if the stance of the \textsc{opinion} matches the stance of the \textsc{source}, e.g., a GW-agree \textsc{opinion} is ascribed to a GW-activist.

\thispagestyle{3303}
Surprisingly, among the 4.3K \textsc{opinions} ascribed to a named entity from the Wikipedia lists, we find that 37\% and 38\% are unfaithfully ascribed in LL and RL, respectively, suggesting that both sides 
frequently attribute \textsc{opinions} to entities that differ from the well-established public positions of those entities.
(See Appendix Tab. \ref{tab:traitor_quotes} for examples of unfaithfully ascribed \textsc{opinions}.)

When we examine the unfaithful instances from LL more closely, we notice that the most frequent \textsc{sources} have ties to the fossil fuel industry (e.g., \textit{Exxon knew that the result of burning fossil fuels would create a climate crisis}), emphasizing the narrative of hypocritical oil companies that have long known about the harmful effects of greenhouse gases. In RL, by contrast, the unfaithful instances quote from a wide-range of activists and scientific bodies, but similarly emphasize these entities' hypocrisy: \textit{Gore admits that carbon dioxide is only responsible for about 40 \% of the warming; NASA concedes that its temperature data are less than reliable}). 

Finally, we ask whether certain \textsc{predicates} are favored for ascribing \textsc{opinions} unfaithfully. We might expect verbs like \textit{admit} and \textit{acknowledge}, which have connotations of reluctance, to be used for this purpose, and for verbs like \textit{declare} and \textit{insist} to be disfavored\textemdash it would be counter-intuitive for a reader of \textit{The New York Times} to see the sentence, \textit{Exxon \textbf{insists} that fossil fuels cause global warming}, for example.

To answer this question empirically, we measure each \textsc{predicate's} tendency to ascribe an \textsc{opinion} to a \textsc{source} with an activist vs. skeptic stance, similar to how we measured \textsc{predicates'} tendency to embed an \textsc{opinion} with a given stance. We retrieve in Tab. \ref{tab:traitor_preds} the \textsc{predicates} that are biased under this measure toward ascribing GW-agree \textsc{opinions} to GW-skeptic \textsc{sources}, and vice versa.

\begin{table}[]
    \centering
    \begin{tabular}{c c}
         Left-leaning media & Right-leaning media \\ \toprule
         understand, \textbf{concede},&realize, \textbf{recall},\\ 
         \textbf{recall}, demonstrate,&learn, see\\ 
         know, acknowledge,& admit, \textbf{concede}\\ 
         agree &reveal\\ \bottomrule
    \end{tabular}
    \caption{\textsc{Predicates} biased toward hypocritical opinion attribution, i.e., attributing an own-side \textsc{opinion} to an opposing-side \textsc{source}, in LL and RL. \textbf{Bolded} \textsc{predicates} are used for hypocritical attribution in both LL and RL.}
    \label{tab:traitor_preds}
\end{table}
\thispagestyle{3304}
Interestingly, in addition to verbs we expected (\textit{acknowledge, admit, concede}), we also find verbs like \textit{understand, agree, realize, know}. One tendency among these verbs seems to be that they denote non-spoken acts of belief. Intuitively, it would be incompatible with real world events to describe Exxon as vocally denouncing fossil fuels or Al Gore as vocally criticizing climate science, but it \textit{is} possible to describe such entities as silently holding contradictory beliefs (and in doing so, highlight their hypocrisy). However, we also see exceptions (\textit{demonstrate} in LL, \textit{reveal} in RL), suggesting that more complex interactions are involved.

\section{Discussion and future work}

In this work, we introduced \textbf{GWSD}, a novel dataset of 2K sentences from news media for studying GW stance. Using our dataset, we trained a weighted BERT model competitive with human performance to predict the stance of 500K opinions in news articles.
Our initial analyses showed that both sides of the GW debate make use of framing devices in largely symmetric ways, though GW-skeptic media exhibits more opponent-doubt, in line with prior work on the propagation of GW skepticism \citep{oreskes2011merchants}. We also found that both sides exhibit considerable amounts of unfaithful opinion attribution, in particular to portray figures as hypocritical. Future work could take a more fine-grained approach to our analyses, such as disaggregating op-ed articles from non-op-eds or adopting labels for outlet stance beyond the binary ``right-'' vs. ``left-leaning.'' We also categorized named entities as either activists or skeptics, which obscures distinctions between, e.g., corporations with economic incentives for GW skepticism vs. individuals that may be ideologically motivated. 

Our methodology may also be useful for work in argument mining: the main object of our inquiry\textemdash ascribed \textsc{opinions} and the linguistic devices of \textsc{source} and \textsc{predicate} used as syntactic markers of the attributive act\textemdash represents a novel dimension along which to analyze how premises are used to support claims \citep{stabgurevych17}. 

Our work also highlights challenges inherent to studying stance: we found that many items can be ambiguous at the sentence-level, without a single ``true'' stance, and that demographic attributes like party affiliation and gender can affect how people respond. At the same time, we showed how Bayesian modeling can be used to account for this variation. Such findings reinforce the idea that NLP should be conscious of who the training data comes from, and how a model might be biased as a result. We hope that future research can benefit from and extend the current work to study argumentation inclusive of the many subjective and demographically-diverse attitudes in our society.

\section*{Acknowledgements}
We thank the reviewers and the Stanford NLP Group for helpful feedback, and Adina Abeles for feedback on the demographics portion of the MTurk task.

\newpage
\bibliographystyle{acl_natbib}
\begingroup\thispagestyle{3305}{\fancyhf{}\setcounter{page}{3305}\fancyfoot[C]{\normalsize{\thepage}}}
\bibliography{emnlp2020}
\endgroup

\clearpage

\section*{Appendices}

\appendix
\thispagestyle{3308}
\section{Data collection details}
\label{sec:data collection details}
\paragraph{URL filters} 
We filtered out articles that may be irrelevant on the basis of containing one of the following URL tags:

/automobiles/, /autoreviews/, /autoshow/,
    /business/, /campaign\text{-}stops/, /crosswords/,
    /booming/, /giving/, /gmcvb/, /jobs/, /lens/,
    /letters/, /newyorktoday/, /nutrition/, /sept-11-reckoning/, /smallbusiness/, /sunday-review/, /garden/, /arts/, /theater/, /sports/, /dining/, /books/, /weekinreview/, /your-money/, /movies/, /fashion/, /technology/, /pageoneplus/, /travel/, /nytnow/, /public-editor/, /education/, /learning/, /podcasts/, /style/, /t-magazine/, /reader-center/, /awardsseason/, /briefing/, /dealbook/, /es/, /greathomesanddestinations/, /interactive/, /media/, /mutfund/, /obituaries/, /personaltech/, /realestate/, /smarter-living/, /todayspaper/, /your-money/, /yourtaxes/, /slideshow/, /interactive/, /tag/, /author/, /clips/, /podcasts/, /subject/, /authors/, /category/, /person/, /category/, /shows/, /video/, /topic/, /topics/, /de/, /tags/, /slideshow/, /interactive/, /transcripts/, /headlines/

\paragraph{Article deduplication details}
\label{sec:appendix title dedup}
We deduplicated articles by normalized URLs. In addition, we noticed that the same article corresponded in some cases to multiple different normalized URLs in our dataset, due to hyperlinking from different sections of a news site (e.g., blog section, RSS feed, front page). We de-duplicated these articles by comparing the article titles using a criterion adapted from Petersen et al. \citeyearpar{petersen2019discrepancy}: for two titles $T_j, T_k$ with Damerau-Levenshtein edit distance of $D_{jk}$, if 
\begin{equation*}
    D_{jk} \leq 0.2 \cdot Min(|T_j|,|T_k|),
\end{equation*}
then we consider the two titles, and hence the corresponding URLs, to index the same article.

\section{\textsc{Source, Predicate, Opinion} extraction algorithm}
\label{sec:appendix spacy pipe}

\begin{enumerate}
    \item Find complement clause(s) in the dependency parse of a sentence, i.e., sub-tree(s) whose root has the dependency label ``ccomp'' (= \textsc{opinion});
    \item Get head(s) of the complement clause(s), which correspond to the main verb that syntactically embeds the comp. clause (= \textsc{predicate}); get children of the \textsc{predicate} with the dep. label ``prt'' (particle) in cases of multi-token verbs, e.g. \textit{point out};
    \item To find the \textsc{source}, first check if the \textsc{predicate} token is a participle (e.g., ``a researcher, \textbf{warning} that [...]''\textemdash if yes, then find the head noun, otherwise, look within all children of \textsc{predicate} and find the syntactic subject (token with the ``nsubj*'' dependency label). In some cases, the head noun/syntactic subject may have the dependency label ``relcl'', indicating that it's inside a relative clause (e.g., ``[...], \textbf{who} warns that'')\textemdash in this case, the true \textsc{source} is the antecedent of the relative pronoun, which we fetch by getting the head of the relative pronoun;
    \item Get additional modifiers of \textsc{source, predicate} and \textsc{opinion} by recursively retrieving their children.
\end{enumerate}

\section{Lexical filters} \label{sec:appendix lexical filters}

We use the following lexical resources to filter extracted complement clauses to true indirect statements on the topic of GW.
\paragraph{The Indiana Lists}

Our algorithm returns subjunctive clausal complements, e.g., ``Politicians require that [oil companies pay a carbon tax]'', which are nearly identical to embedded opinions, e.g., ``Politicians claim that [oil companies pay a carbon tax]''. The Indiana Lists \citep{alexander1964some,bridgeman1965more} categorize predicates according to whether they syntactically embed a subjunctive or indicative complement clause. We keep extracted (\textsc{source}, \textsc{predicate}, \textsc{opinion}) tuples only if the \textsc{predicate} lemma is one of 418 indicative-clause-embedding verbs in these lists. This filter also effectively excludes extracted instances such as ``We watch [oil companies pay a carbon tax]''. 

\paragraph{Implicatives}
In addition to separating (\textsc{S,P,O}) tuples  with overt negations (\textit{The researchers did \textbf{not} say [that the effects of global warming are clear]}, \textit{\textbf{No} studies find that [...]}), we also need to separate tuples that are implicitly negated (\textit{The studies \textbf{fail to} find that [...], \textit{Researchers \textbf{refuse to} say [...]}} in order to accurately study how opinions are attributed. Since the dependency parser only recognizes explicit cases of negation, we use a list of 92 implicative constructions from Cases et al. \citeyearpar{cases-etal-2019-recursive} to exclude tuples where the \textsc{predicate} is in the scope of such an implicitly negating expression.

\paragraph{Indirect questions} We exclude complement clauses that represent indirect questions (\textit{Scientists ask \textbf{what the future of nuclear looks like}}) by excluding tuples that have a question word from the set \{\textit{who, what, when, where, how, whether, which}\} as the complementizer. 

\paragraph{Topic keywords}\label{app:prec-keywords}

climat, climact, global, warm, carbon, fossil, oil, energi, environ, co2, green, ice, glacier, glacial, melt, sea, temperatur, heat, hot, methan, greenhous, arctic, antarct, celsiu, fahrenheit, ecosystem, pole, environ, coal, natur, human, economi, electr, futur, health, scienc, econom, air, pollut, fire, wildfir, ipcc, epa, market, scientist, earth, planet, wind, solar, record, fuel, ocean, nuclear, scientif, pipelin, emit, emiss, concensu, renew, accord, forest, pruitt, drought, hurrican, atmospher, activist, coast, agricultur, water, plant, weather, polar

\section{AMT task details} 
\label{app:appendix example mturk}
\thispagestyle{3309}
To choose the subset of items for annotations from our full set of extracted \textsc{opinion} spans, we filter to items that contain a smaller set of keywords (``climate'', ``warming'', ``carbon'', ``co2'', or ``fossil fuels'') and make a manual selection for each round of annotation such that the final sample is roughly balanced across different outlets. 

We settle on a task design as follows: annotators are told that we are collecting their judgments of GW stance for a series of sentences; we then show an instructions page and guide them through 6 practice trials. They then annotate the main trial items for \textit{agreeing}, \textit{disagreeing}, or being \textit{neutral} with respect to the target opinion, ``Climate change/global warming is a serious concern.'' Additional help text for each label is adapted from the set-up described in Mohammad et al. \citeyearpar{mohammad2016semeval}. The main trial items consist of 5 screen sentences and 30-50 sentences that have been transformed from the extracted \textsc{opinion} using basic operations such as cleaning whitespace, capitalizing the first word, adding clause-final punctuation, matching for tense, and substituting abbreviations of named entities with the non-abbreviated form. 

\begin{figure}
    \centering 
    \includegraphics[scale=0.34]{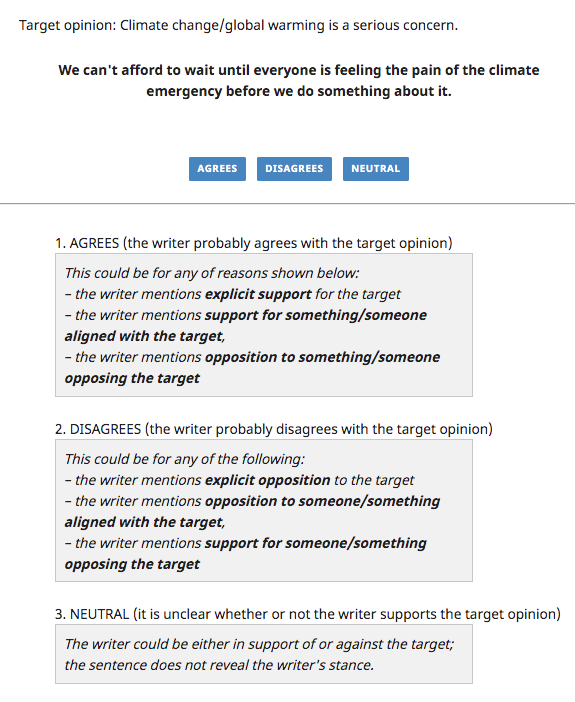}
    \label{fig:my_label}
\end{figure}

We divide the annotation into 5 rounds and recruit 8 annotators to annotate each item. Other than one worker who did the task on 3 different rounds, all other annotations come from unique annotators. We also restrict to annotators whose IP address is in the US, who have a minimum HIT approval rating of 98\%, and at least 1,000 HITs approved. We collect annotator age, gender, level of education, political affiliation, state of residence, as well as measures of their own stance towards GW borrowed from the American Public Opinion on Global Warming project.\footnote{https://pprggw.wordpress.com/} There is some demographic imbalance in our total sample of annotators (see Tab. \ref{tab:dem_stats} in \ref{app:demographic}) but the distribution is similar to the estimated demographics of the AMT population located in the US as a whole \citep{ross2009turkers}. The price per item was set to ensure that workers were paid the California minimum wage of \$12 USD per hour.

\section{Demographic and linguistic effects on annotations}
\label{app:demographic}

The marginal statistics for annotator demographics are given in Table \ref{tab:dem_stats}, and show a relatively representative sample in terms of age, gender, education, and political affiliation, though women are are distinctly under-represented.\footnote{For political affiliation by age and gender in the US, see \url{http://pewrsr.ch/2FVWtww}}

\begin{table}[ht]
    \centering
    \begin{tabular}{l r}
        Answer & \% of annotators \\
        \hline
        Age over 34 & 48.3 \% \\
        Female & 37.3 \% \\
        Male & 62.5 \% \\
        College degree or higher  & 66.5\% \\
        Democrat & 46.0 \% \\
        Republican & 21.2 \% \\
        Independent & 28.8 \% \\
        Other political affiliation & 4.0 \%
        \end{tabular}
    \caption{Demographic information on  the 400 Mechanical Turk annotators who participated in our study.}
    \label{tab:dem_stats}
\end{table}

In order to measure the bias associated with various characteristics of annotator demographics, we make use of the hierarchical ordinal logistic model given below. In this model, $Y_{ij}$ is the response of annotator $j$ to instance $i$ (taking a value in $\{1,2,3\}$, corresponding to ``disagree'', ``neutral'', ``agree''). In addition, $q_i$ is the unnormalized stance associated with instance  $i$ (on a spectrum from ``disagree'' to ``agree''), $w_j$ is the bias associated with worker $j$, $X_j$ is a vector of covariates associated with worker $j$, $\sigma_q^2$ and $\sigma_w^2$ are learned variance parameters, and $c_1$ and $c_2$ are learned thresholds. We model the probability of each response according to:
\begin{equation} 
  p(Y_{ij} = k) = 
    \begin{cases}
      1 - g(\eta_{ij} - c_1) & \text{if $k = 1$}\\
      g(\eta_{ij} - c_1) \\
      \quad -~g(\eta_{ij} - c_2) & \text{if $1 < k < K$}\\
      g(\eta_{ij} - c_2) & \text{if $k = K$} \\
    \end{cases} 
\end{equation}
where
\begin{align}
  \eta_{ij} &= q_i + w_j \\
  q_i &\sim \mathcal{N}(0, \sigma_q^2) \\
  w_j &\sim \mathcal{N}(\beta^T X_j, \sigma_w^2) 
\end{align}

To complete the model, we place weakly informative half-normal priors on $\sigma_q^2$ and $\sigma_w^2$, and weakly informative normal priors on $\beta$.

\thispagestyle{3310}
Using the above specification, we fit a series of models in which $X_j$ represents, in turn, each of the covariates individually, followed by a series of combined models. We fit these models in Stan using 5 chains with 2000 samples, the first half thrown away as burn in. 

Table \ref{tab:agree_table} shows the estimated effects from each model on the propensity to respond with ``agree'' relative to ``neutral''. Those with 95\% credible intervals which exclude 1.0 are marked in bold. The results on the propensity to respond with ``neutral'' relative ``disagree'' are not shown, but are broadly similar.

\begin{table*}[ht]
    \centering
    \begin{tabular}{l ccccccccccc}
        Covariate & M1 & M2 & M3  & M4 & M5 & M6 & M7 & M8\\
        \hline
        Age over 34 & 0.98 &&&&&&& 0.98  \\
        Female & & \bf{1.04} &&&&& \bf{1.04} & \bf{1.04}  \\
        College degree or higher &&& 1.0 & &&&& 1.0 \\
        Democrat &&&& \bf{0.96}&& 0.97 & 0.98 & 0.98 \\
        Republican &&&&& \bf{1.06} & \bf{1.05} & \bf{1.04} & \bf{1.05} \\
    \end{tabular}
    \caption{Effects of annotator demographics on the propensity to respond with ``agree'' rather than ``neutral''. Coefficients in bold have 90\% credible intervals which exclude 1.0.}
    \label{tab:agree_table}
\end{table*}

In addition, we test whether the political leaning of the source outlet has any effect on the annotations received by the items drawn from the source. We find, unsurprisingly, that items from left-leaning media (LL) are significantly more likely to receive ratings of ``agree'', but we do not find a significant difference in level of annotator agreement for items drawn from LL vs. RL. Finally, we find that item length (no. of words) is slightly correlated with IAA (measured as entropy over labels; Spearman's $\rho$ = 0.06, p = 0.016).

\section{Estimating Stance Distributions}
\label{app:aggregating}

To aggregate all ratings and obtain estimates of the stance distribution for each instance, we use a variant of the above model which allows inferring a distribution for each instance and each worker, along with a parameter representing the degree to which an annotator is failing to pay attention to the instance being annotated. Although the ``disagree'', ``neutral'', and ``agree'' categories can be treated as ordered (as above), here we treat them as unordered nominal categories, so as to allow for the possibility, for example, that an instance evokes both ``agree'' and ``disagree'', but not ``neutral'' (i.e. it is ambiguous, but clearly not neutral). 

Let $Y_{ij}$ be the response from worker $j$ to item $i$, let $q_{ik}$ be the degree to which label $k$ applies to instance $i$, and let $w_{jk}$ be the bias of worker $j$ towards label $k$. Finally, let $v_j$ be the vigilance of worker $j$ (i.e. the degree to which they pay attention to the prompt). We assume the following model
\begin{align}
  Y_{ij} &\sim \text{Multinomial}(\text{Softmax}_k(\eta_{ij})) \\
  \eta_{ijk} &= v_j \cdot q_{ik} + (1-v_j) \cdot w_{jk} \\
  q_{ik} &\sim \mathcal{N}(\mu_k, \sigma_q^2) \\
  w_{jk} &\sim \mathcal{N}(0, \sigma_w^2) 
\end{align}
and fit it in Stan, placing weakly informative priors on $\sigma_q^2$, $\sigma_w^2$, and a uniform prior on $v_j \in [0, 1]$. In order to help stabilize the model we set the mean parameter of the prior on $q_{ik}$ to be $\mu_k = \log{p_k}$, where $p_k$  is the overall proportion of the corresponding response in the data.

\begin{figure}
    \centering
    \includegraphics[scale=0.6]{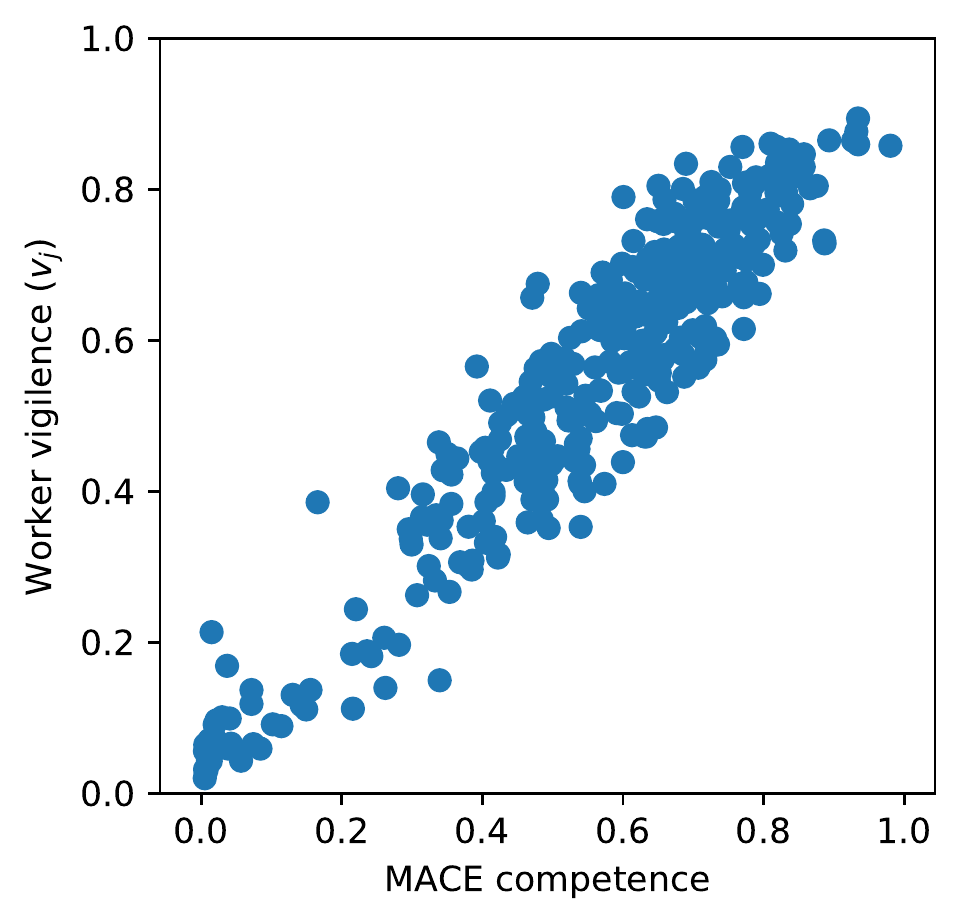}
    \caption{Showing the correlation between worker competence (estimated using MACE) and worker vigilance (estimated using our model) for the 400 annotators who participated in our data collection.}
    \label{fig:mace_comp}
\end{figure}

\begin{table}[]
    \centering
    \begin{tabular}{l|c|c|c|}
        MACE / Ours & disagree & neutral & agree \\
        \hline
        disagree & 386 & 6 & 0 \\
        \hline
        neutral & 12 & 852 & 19 \\
        \hline
        agree & 2 & 15 & 785 \\
        \hline
    \end{tabular}
    \caption{Confusion matrix of (dis)agreements between MACE and our model}
    \label{tab:mace_cfm}
\end{table}

\begin{figure}
    \centering
    \includegraphics[scale=0.55]{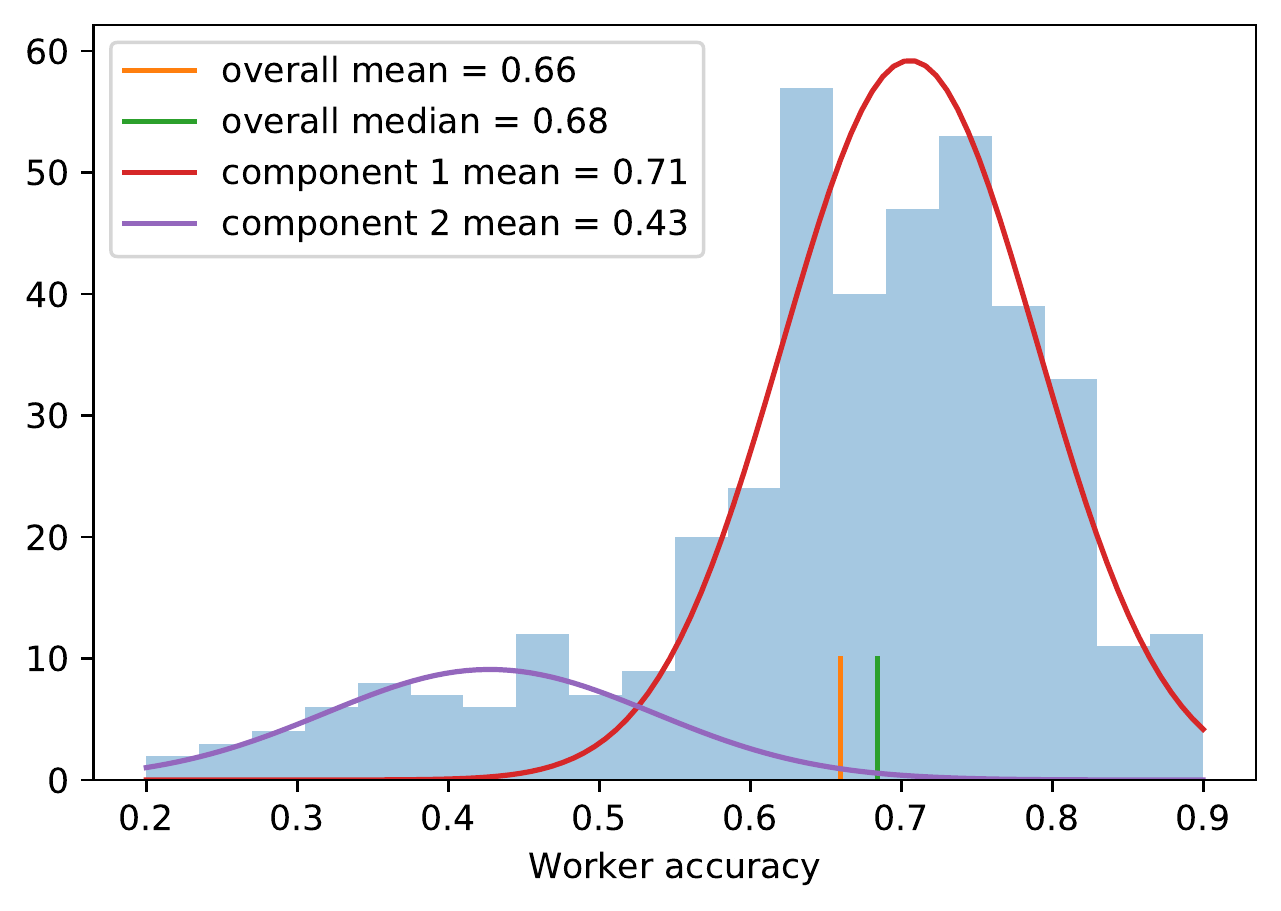}
    \caption{To estimate human performance, we refit the label aggregation model multiple times, each time leaving out 10\% of annotators, and then comparing their annotations against the inferred label for the corresponding items. This plot shows that human performance appears to be a mixture of two distributions representing low and high agreement annotators.}
    \label{fig:human}
\end{figure}

In order to estimate human performance for the purpose of comparison, we fit this model multiple times, but each time leave out a random 10\% of the annotators. As can be seen in Figure \ref{fig:human}, there is great variation in the degree to which annotators agree with the label inferred from the remaining 90\% of annotators. To characterize the distribution of work accuracies, we fit a mixture of two normal distributions, and report the mean of the distribution corresponding to the high-accuracy annotators in the main paper (0.71).

\section{Additional BERT experiments}
Leading up to our hyperparameter search and baseline comparison, we experiment with a variety of training set-ups. Due to class imbalance in our training data (see Table \ref{tab:label_stats}), we try downsampling the majority classes as well as upsampling the minority class by adding back translations thereof. However, we did not obtain performance gains from either strategy in preliminary experiments. We further experiment with additional features in the form of the pre- and postceding \textit{n} sentences ($n = [1,2]$) surrounding a training example, but did not obtain performance gains. We are also limited in the kinds of additional features (e.g., the political leaning of the outlet that a sentence comes from, the source entity that a sentence is attributed to) that we can use, since our goal is to analyze how the stance of the embedded statement is correlated with precisely these variables. We also try fine-tuning BERT first on a language modeling task (using our raw news data) and a natural language inference task (using the SNLI \citep{bowman-etal-2015-large} and MNLI \citep{williams-etal-2018-broad} datasets), respectively, prior to fine-tuning for sequence classification, but obtain no performance gains. Finally, we experiment with using tweets from known GW activists/skeptics as well as GW article headlines taken from extreme liberal/conservative news sources as additional training data, inferring labels based on the stance of the Twitter user or news source. However, we find that adding these examples yields a lower performance compared to using only the human-annotated data. This is not too surprising, given that embedded statements in the news tend to be longer and more complex than tweets/headlines.

\thispagestyle{3311}
\section{Hyperparameter Tuning}
\label{app:hyper}
All experiments took less than 7 days on one GPU (16 cores, 2.6GhZ, 128GB mem). For the BERT-based models (110M parameters), we use a maximum sequence length of 256, a batch size of 16, and train for 7 epochs, saving a checkpoint after each epoch. In addition, we perform a grid search over the following hyperparmaters:
\begin{itemize}
    \item Label weights: [True, False]
    \item Language model fine-tuning: [True, False]
    \item Target opinion as second input: [True, False]    
    \item Learning rate: [1e-5, 2e-5, 4e-5]
\end{itemize}

We train models for each combination of settings using five random seeds, and ultimately choose the hyperparmeter configuration (including number of training epochs and random seed) that has the best validation performance, averaged over five folds, for a total of 600 configuration tested (including seeds and folds). We then retrain a model using those hyperparameter values on all non-test data.

Because we are using grid search, we can conveniently compare the effects of various hyperparameter choices. The overall average validation performance was 0.71, with a standard deviation of 0.04, and a 95\% interval of [0.64, 0.77]. 
Table \ref{tab:bert_hyperparams} shows the average increase in accuracy associated with each hyperparameter choice, along with a $p$-value computed using a Wilcoxon signed-rank test. As can be seen, using label weights leads to a significant increase in accuracy, as does using a learning rate of 2e-5 in comparison to 1e-5.

\begin{table}[]
    \centering
    \begin{tabular}{l r r}
        Hyperparameter & $\Delta$ accuracy & $p$-value \\
        \hline
        Label weights & 0.020 & $< 0.001$ \\
        LM fine-tuning & 0.004 & 0.11 \\
        Target opinion & -0.002 & 0.48 \\
        LR 2e-5 vs 1e-5 & 0.009 & 0.03 \\
        LR 4e-5 vs 1e-5 & 0.002 & 0.35
    \end{tabular}
    \caption{Estimated effects of various hyperparameter choices on the average validation performance of the $\textrm{BERT}_{base}$ model. $p$-values are obtained using a Wilcoxon signed-rank test on the paired results from grid search.}
    \label{tab:bert_hyperparams}
\end{table}

\thispagestyle{3312}
For the linear models (91504 params), we consider both logistic regression and SVM models, again using grid search and choosing the best-performing model on average validation performance, as described above. For the SVM, we search over all combinations of the following hyperparameters:
\begin{itemize}
    \item Label weights: [True, False]
    \item n-gram order: [1, 2]
    \item kernel [rbf, linear, polynomial]
    \item gamma [scale, auto]
    \item Stopword removal: [True, False]
    \item Convert digits: [True, False]
    \item Regularization strength \{0.01, ..., 1000\}
\end{itemize}

For the logistic regression model, we search over all combinations of the following hyperparameters:
\begin{itemize}
    \item Label weights: [True, False]
    \item n-gram order: [1, 2]
    \item Stopword removal: [True, False]
    \item Convert digits: [True, False]
    \item Regularization type: [$l_1$, $l_2$]
    \item Regularization strength \{0.01, ..., 1000\}
\end{itemize}

The mean validation accuracy among a total of 640 linear models tested is 0.56, with a standard deviation of 0.06, and a (0.41-0.62) 95\% confidence interval. The linear model which performed best on validation data was a logistic regression bigram model using label weights, trained with $l_2$ regularization, no stopword removal, no digit conversion, and regularization strength of 1.0.

\begin{figure}
    \centering
    \includegraphics[scale=0.55]{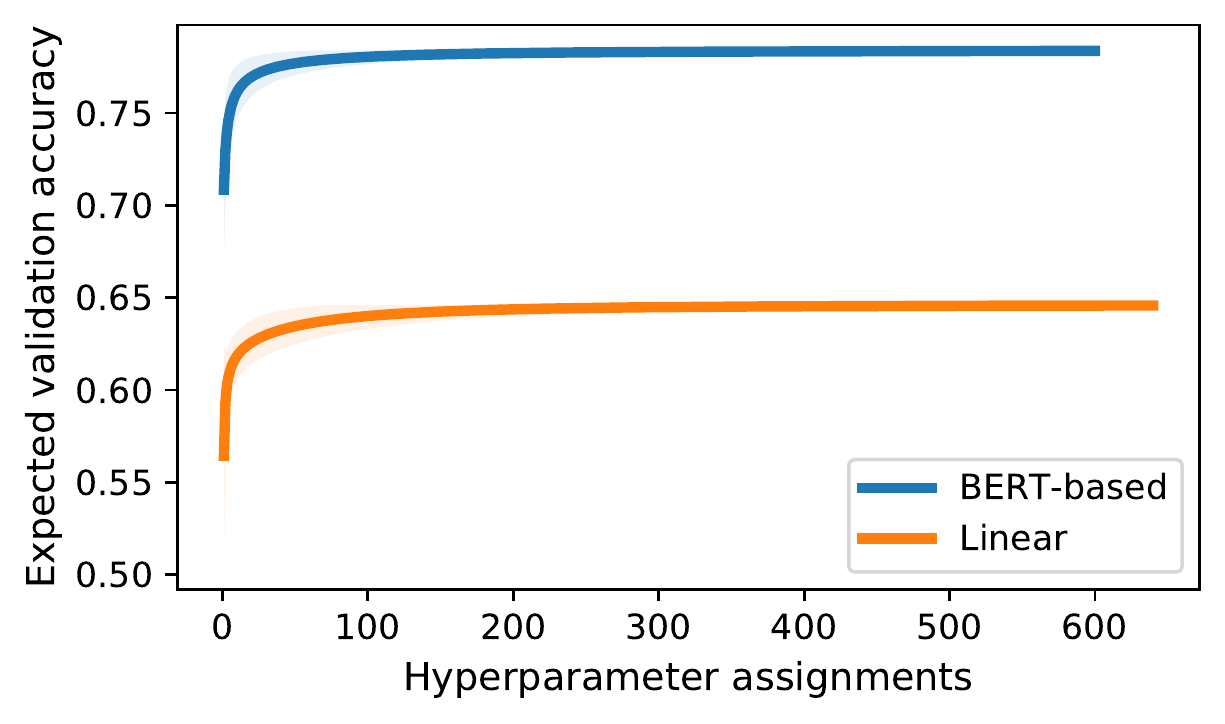}
    \caption{Expected validation performance of both types of models using validation accuracy scores from the hyperparameter grid search.}
    \label{fig:expected_val}
\end{figure}

Figure \ref{fig:expected_val} compares these results directly, showing that the expected validation performance \citep{dodge.2019} of the BERT-based models is uniformly better than that of the linear models, at least in terms of number of hyperparameter assignments.

\section{Framing devices}
\label{app:framing devices}
\paragraph{Affirming devices}
\begin{itemize}
    \item \textbf{Factive and semi-factive verbs:} uncover, realize, know, understand, learn, concede, remember, recall, discover, show, reveal, see, forget, find, point out, indicate, acknowledge, admit, realize, notice
    \item \textbf{High-commitment verbs:} certify, verify, corroborate, affirm, confirm, agree, conclude
    \item \textbf{High commitment adjectives:} proven, settled, conclusive, definitive
    \item \textbf{Hyping adjectives:} famed, unequivocal, skilful, notable, strong, famous, Nobel, skillful, Nobelist, Nobel Laureate, Nobel prize winner, Nobel prize winning, prize winning, award winning, distinguished, well-grounded, esteemed, proficient, key, evidence, noted, top, preeminent, breakthrough, significant, intelligent, of import, celebrated, novel, recent, major, landmark, important, distinguished, renowned, peer-reviewed, expert, leading
    \item \textbf{Consensus of evidence adjectives:} thousand, 1000, hundred, 100, unanimous, diverse, substantial, many, multiple, dozen, numerous
\end{itemize}

\paragraph{Doubting devices}
\begin{itemize}
    \item \textbf{Neg-factive verbs:} pretend, lie, claim, allege, assume
    \item \textbf{Low commitment verbs:} doubt, dispute, debate
    \item \textbf{Argumentative verbs:} boast, declare, argue, maintain, contend, insist, proclaim, assert, brag, tout, convince
    \item \textbf{Low commitment modifiers:} narrative, evangelical, hoax, dubious, alleged, in question, so-called
    \item \textbf{Undermining adjectives:} discredited, debunked, distorted, misleading, inaccurate, corrupted, sketchy, faulty, erroneous, deficient, wrong, flawed, imprecise, incomplete, insufficient, invalid, unreliable, adulterated, false, mistaken, cherry-picked, defective, presumptive, non-peer-reviewed, exaggerated, overdone, overstated, delusive, awry, fake, bad, misguided, substandard, fictive, fictitious, uncomplete, blemished, uncompleted, shoddy, dubitable, lacking, moot, untrue,  problematic, faux, incorrect, inferior
    \item \textbf{Lack of consensus adjectives:} controversial, contentious, debated, few, debatable, contested
\end{itemize}

\thispagestyle{3313}
\section{Quantifying bias toward framing \textsc{opinions} with a GW-agree stance}
\label{app:log odds}

We measure, $B_{f,L}$, the tendency for a framing device, \textit{f}, within media with leaning $L$, to frame a GW-agree \textsc{opinion} as:
\begin{equation*}
    B_{f,L} = \log \left( {\frac{a_{f}}{A-a_f}} \right) - \log \left( {\frac{d_{f}}{D-d_f}} \right),
\end{equation*}
where $a_f$ is the number of times $f$ occurs with a GW-agree \textsc{opinion}, $A$ is the total number of GW-agree \textsc{opinions}, $d_f$ is the number of times $f$ occurs with a GW-disagree \textsc{opinion}, and $D$ is the total number of GW-disagree \textsc{opinions}, all within $L$.

\section{Named entity fuzzy matching}
\label{app: fuzzy matching}
We use FuzzyWuzzy (\url{https://github.com/seatgeek/fuzzywuzzy}) to retrieve fuzzy matches of named entity \textsc{sources}, setting the limit of matches to N = 100. We then manually filter out incorrect matches.

\begin{table*}[ht]
    \centering
    \begin{tabular}{l} \toprule
         \tcbox[hbox,tcbox raise base,enhanced,fontupper=\small\bfseries,drop fuzzy shadow southwest,size=fbox,
    colframe=red!20!black,colback=red!20,nobeforeafter]{Tillerson} \textbf{acknowledged} that \begin{tcolorbox}[tcbox raise base,enhanced,fontupper=\small\bfseries,drop fuzzy shadow southwest,size=fbox,width=3.1in,
    colframe=red!20!black,colback=blue!20,nobeforeafter]
    climate change has `real' and `serious' risks but has previously downplayed climate change effects.
    \end{tcolorbox} \\
    \tcbox[hbox,tcbox raise base,enhanced,fontupper=\small\bfseries,drop fuzzy shadow southwest,size=fbox,
    colframe=red!20!black,colback=red!20,nobeforeafter]{Exxon} \textbf{knows} that \begin{tcolorbox}[enhanced,tcbox raise base,fontupper=\small\bfseries,drop fuzzy shadow southwest,size=fbox,width=3.1in,
    colframe=red!20!black,colback=blue!20,nobeforeafter]
    fossil fuels caused global warming in the 1970s.
    \end{tcolorbox} \\
    \tcbox[hbox,tcbox raise base,enhanced,fontupper=\small\bfseries,drop fuzzy shadow southwest,size=fbox,
    colframe=red!20!black,colback=red!20,nobeforeafter]{Exxon} \textbf{knew} that \begin{tcolorbox}[tcbox raise base,hbox,enhanced,fontupper=\small\bfseries,drop fuzzy shadow southwest,size=fbox,width=3.1in,
    colframe=red!20!black,colback=blue!20,nobeforeafter]
    the result of burning fossil fuels would create a climate crisis.
    \end{tcolorbox} \\ \midrule
    \tcbox[tcbox raise base,hbox,enhanced,fontupper=\small\bfseries,drop fuzzy shadow southwest,size=fbox,
    colframe=red!20!black,colback=blue!20,nobeforeafter]{Gore} \textbf{admits} that \begin{tcolorbox}[hbox,tcbox raise base,enhanced,fontupper=\small\bfseries,drop fuzzy shadow southwest,size=fbox,width=3.1in,
    colframe=red!20!black,colback=red!20,nobeforeafter]
    carbon dioxide is only responsible for about 40 percent of the warming.
    \end{tcolorbox} \\
    Even the \tcbox[hbox,tcbox raise base,enhanced,fontupper=\small\bfseries,drop fuzzy shadow southwest,size=fbox,
    colframe=red!20!black,colback=blue!20,nobeforeafter]{IPCC} \textbf{acknowledges} that \begin{tcolorbox}[enhanced,tcbox raise base,fontupper=\small\bfseries,drop fuzzy shadow southwest,size=fbox,width=3.1in,
    colframe=red!20!black,colback=red!20,nobeforeafter]
    their previous estimates of “ climate sensitivity ” to greenhouse gases their reported in 2007 were significantly exaggerated.
    \end{tcolorbox} \\
    \tcbox[hbox,tcbox raise base,enhanced,fontupper=\small\bfseries,drop fuzzy shadow southwest,size=fbox,
    colframe=red!20!black,colback=blue!20,nobeforeafter]{NASA} \textbf{concedes} that  \begin{tcolorbox}[enhanced,tcbox raise base,fontupper=\small\bfseries,drop fuzzy shadow southwest,size=fbox,width=3.1in,
    colframe=red!20!black,colback=red!20,nobeforeafter]
    its temperature data are less than reliable.
    \end{tcolorbox} \\ \bottomrule
    \end{tabular}
    \caption{Examples of unfaithfulness in opinion attribution. \textbf{Top}: Examples of LL attributing \tcbox[hbox,tcbox raise base,enhanced,fontupper=\small\bfseries,drop fuzzy shadow southwest,size=fbox,
    colframe=red!20!black,colback=blue!20,nobeforeafter]{GW-agree \textsc{opinions}} to \tcbox[hbox,tcbox raise base,enhanced,fontupper=\small\bfseries,drop fuzzy shadow southwest,size=fbox,
    colframe=red!20!black,colback=red!20,nobeforeafter]{GW-skeptic \textsc{sources}}. \textbf{Bottom}: Examples of RL attributing \tcbox[hbox,tcbox raise base,enhanced,fontupper=\small\bfseries,drop fuzzy shadow southwest,size=fbox,
    colframe=red!20!black,colback=red!20,nobeforeafter]{GW-disagree \textsc{opinions}} to \tcbox[hbox,tcbox raise base,enhanced,fontupper=\small\bfseries,drop fuzzy shadow southwest,size=fbox,
    colframe=red!20!black,colback=blue!20,nobeforeafter]{GW-activist \textsc{sources}}. The IPCC refers to the U.N.'s Intergovernmental Panel on Climate Change.} 
    \label{tab:traitor_quotes}
\end{table*}

\newpage
\newpage
\thispagestyle{3314}
\section{Results on non-top-5 LL and RL media}
\label{app:subsampled results}

\begin{figure}[ht]
    \centering
    \includegraphics[scale=0.3]{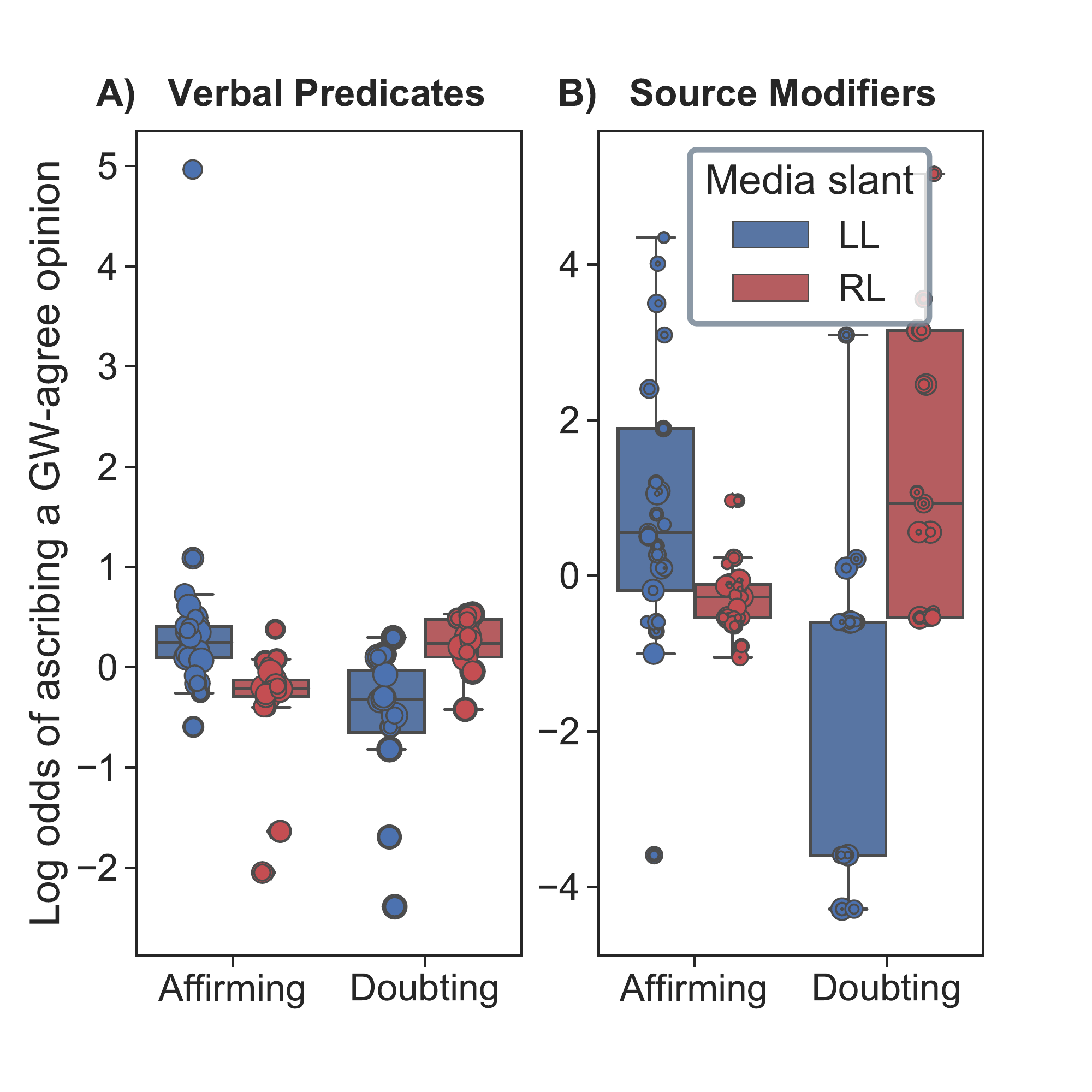}
    \caption{Log odds of ascribing a GW-agree \textsc{opinion} in LL and RL for affirming and doubting \textsc{predicates} (left panel) and \textsc{source} modifiers (right panel).}
    \label{fig:my_label}
\end{figure}

\clearpage
\thispagestyle{3315}
\begin{figure}[ht]
    \centering
    \includegraphics[scale=0.3]{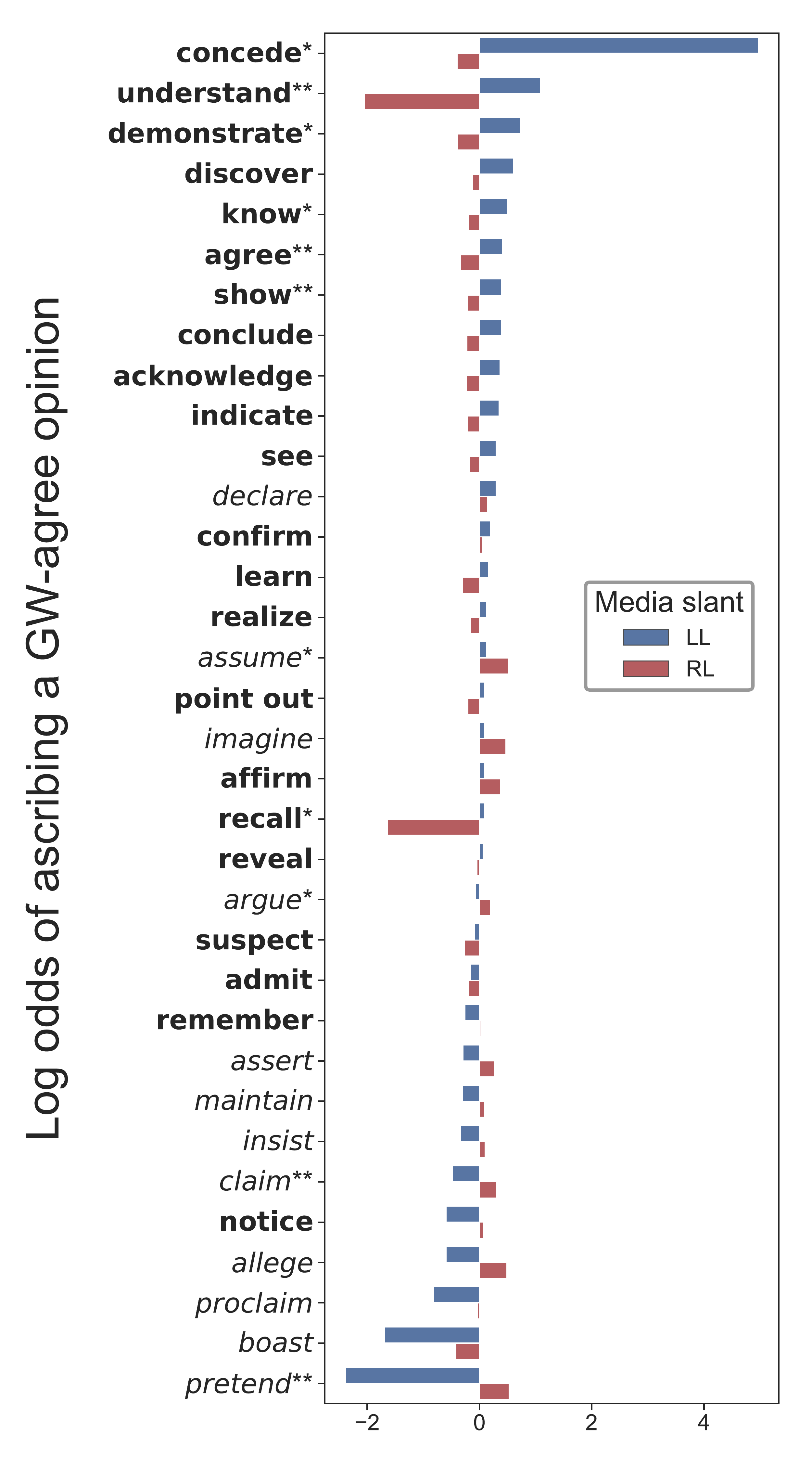}
    \caption{Log odds of ascribing a GW-agree \textsc{opinion} in RL and LL for different \textsc{predicates}, excluding the top 5 outlets by number of articles in each. A double asterisk (**) indicates a significant bias for GW-agree \textsc{opinions} in both LL and RL; (*) indicates significance in one side. Significance ($p<0.05$) is determined via a chi-squared test and applying Benjamini-Hochberg correction with a false discovery rate of 0.1. Word order is given in descending value of log odds, as measured in LL.}
    \label{fig:my_label}
\end{figure}
\begin{figure}[ht]
    \centering
    \includegraphics[scale=0.3]{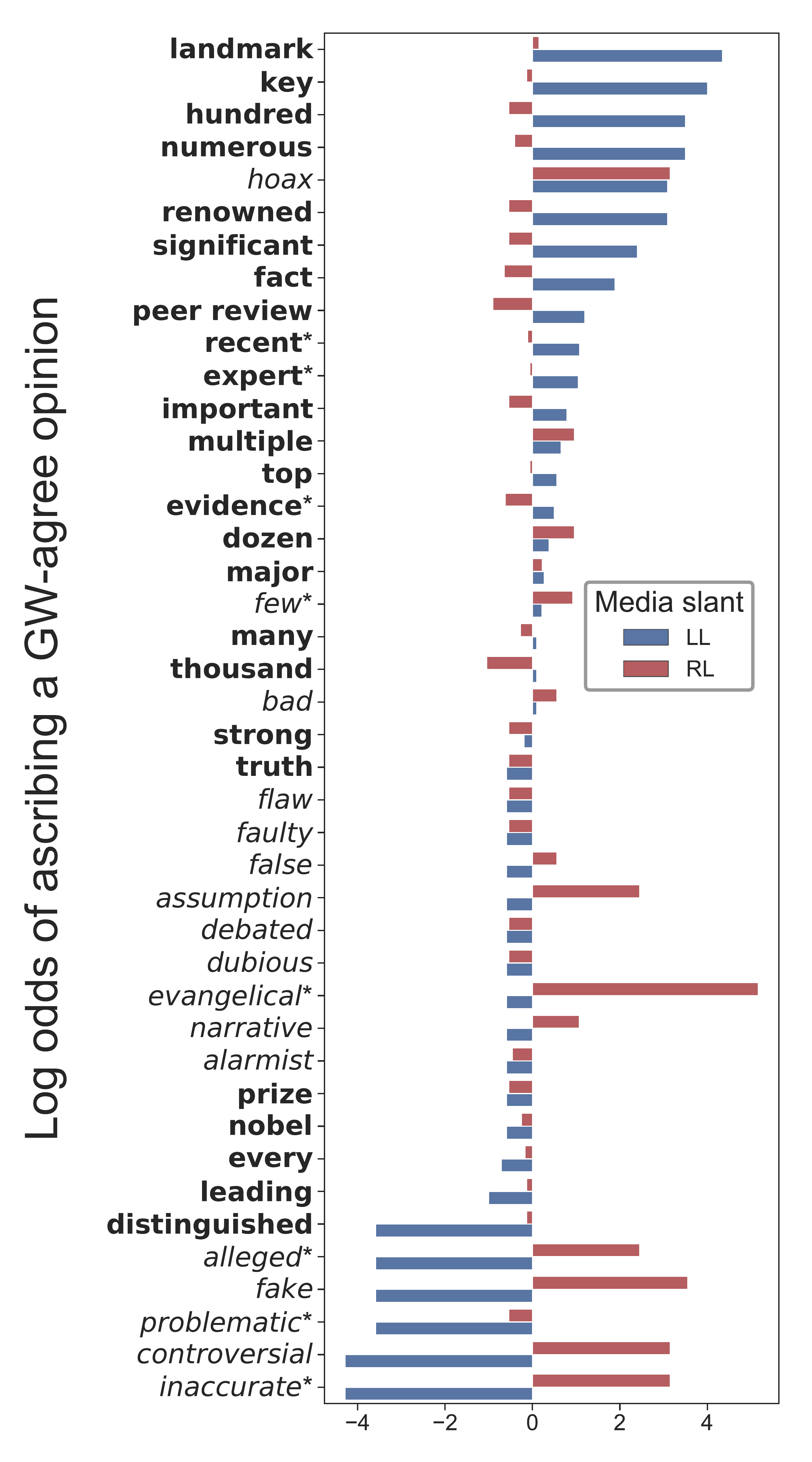}
    \caption{Log odds of ascribing a GW-agree \textsc{opinion} in RL and LL for different \textsc{source} modifiers, excluding the top 5 outlets by number of articles in each. A single asterisk (*) indicates a significant bias for GW-agree \textsc{opinions} in either LL or RL. Significance ($p<0.05$) is determined via a chi-squared test and applying Benjamini-Hochberg correction with a false discovery rate of 0.1. Word order is given in descending value of log odds, as measured in LL.}
    \label{fig:my_label}
\end{figure}

\end{document}